\documentclass[10pt,journal,compsoc]{IEEEtran}
\newif\ifpeerreview

\peerreviewfalse

\usepackage[nocompress]{cite}
\usepackage{url}
\usepackage{amsmath,amssymb,graphicx}
\usepackage[switch]{lineno}
\usepackage{booktabs}
\usepackage{threeparttable}
\usepackage[pagebackref=true,breaklinks=true,colorlinks,bookmarks=false]{hyperref}
\usepackage{tabularx}
\usepackage{xcolor} % for customized colors

\def\eg{\emph{e.g.}}

\def\ie{\emph{i.e.}}

\def\etc{\emph{etc.}}

\def\etal{\emph{et al.}}

\newcommand{\Tref}[1]{Table~\ref{#1}}

\newcommand{\Eref}[1]{Equation~(\ref{#1})}
\newcommand{\Fref}[1]{Figure~\ref{#1}}
\newcommand{\Sref}[1]{Section~\ref{#1}}

\newcommand{\bl}{\mathbf{l}}
\newcommand{\bi}{\mathbf{i}}
\newcommand{\bd}{\mathbf{d}}
\newcommand{\bL}{\mathbf{L}}
\newcommand{\bI}{\mathbf{I}}
\newcommand{\bA}{\mathbf{A}}
\newcommand{\bS}{\mathbf{S}}

\newcommand{\bn}{\mathbf{n}}
\newcommand{\bD}{\mathbf{D}}
\newcommand{\bW}{\mathbf{W}}

\definecolor{orange}{rgb}{1.0, 0.6, 0.0}

\newcolumntype{L}[1]{>{\raggedright\arraybackslash}p{#1}}
\newcolumntype{C}[1]{>{\centering\arraybackslash}p{#1}}
\newcolumntype{R}[1]{>{\raggedleft\arraybackslash}p{#1}}

\newcommand{\paperID}{39}

\title{DeRenderNet: Intrinsic Image Decomposition of Urban Scenes with Shape-(In)dependent Shading Rendering}

\author{Yongjie~Zhu,
        Jiajun~Tang,
        Si~Li,
        and Boxin~Shi,~\IEEEmembership{Senior Member,~IEEE}% <-this % stops a space
\IEEEcompsocitemizethanks{
\IEEEcompsocthanksitem Y. Zhu and S. Li are with the School of Artificial Intelligence, Beijing University of Posts and Telecommunications, China.\protect\\
E-mail: \{zhuyongjie, lisi\}@bupt.edu.cn
\IEEEcompsocthanksitem J. Tang and B. Shi (corresponding author) are with the National Engineering Laboratory for Video Technology, Department of Computer Science and Technology, Peking University, China. B. Shi is also affiliated with the Institute for Artificial Intelligence, Peking University, China.\protect\\
E-mail: \{jiajun.tang, shiboxin\}@pku.edu.cn}% <-this % stops an unwanted space
}

\begin{document}

%--------------------------------------------------------------------------------------------------------

\IEEEtitleabstractindextext{
\begin{abstract}
We propose DeRenderNet, a deep neural network to decompose the albedo and latent lighting, and render shape-(in)dependent shadings, given a single image of an outdoor urban scene, trained in a self-supervised manner. To achieve this goal, we propose to use the albedo maps extracted from scenes in videogames as direct supervision and pre-compute the normal and shadow prior maps based on the depth maps provided as indirect supervision. Compared with state-of-the-art intrinsic image decomposition methods, DeRenderNet produces shadow-free albedo maps with clean details and an accurate prediction of shadows in the shape-independent shading, which is shown to be effective in re-rendering and improving the accuracy of high-level vision tasks for urban scenes.
\end{abstract}

}

% Make Title
\ifpeerreview
\linenumbers \linenumbersep 15pt\relax 
\author{Paper ID \paperID\IEEEcompsocitemizethanks{\IEEEcompsocthanksitem This paper is under review for ICCP 2021 and the PAMI special issue on computational photography. Do not distribute.}}
\markboth{Anonymous ICCP 2021 submission ID \paperID}%
{}
\fi
\maketitle
% \thispagestyle{empty} % camera-ready

%--------------------------------------------------------------------------------------------------------

% The first section title should be wrapped inside a \IEEEraisesectionheading as follows.
\IEEEraisesectionheading{
  \section{Introduction}\label{sec:introduction}
}

%% brief intro to the task of intrinsic decomposition
\IEEEPARstart{I}{ntrinsic} image decomposition aims to decompose an input image into its intrinsic components that separate the
material properties of observed objects from illumination effects~\cite{Barrow1978}. It has been studied extensively that such decomposition results could be beneficial to many computer vision tasks, such as segmentation~\cite{semantic_cvpr19}, recognition~\cite{huang2019msrcnn}, 3D object compositing~\cite{LIME}, relighting~\cite{sfsnet}, \etc~However, decomposing intrinsic components from a single image is a highly ill-posed problem, since the appearance of an object is jointly determined by various factors. An unconstrained decomposition without one or more terms of shape, reflectance, and lighting being fixed, will produce infinitely many solutions.

%% related works in a glance
To address the shape-reflectance-lighting ambiguity, classic intrinsic decomposition methods adopt a simplified formulation by assuming an ideally diffuse reflectance for observed scenes. These methods decompose an observed image as the pixel-wise product of a reflectance map and a greyscale shading image, leaving the lighting color effects on the reflectance image~\cite{bell14intrinsic,zhou2015learning,li2018cgintrinsics,2017reflectanceFiltering}. By learning the reflectance consistency of a time-lapse sequence, the illumination color can be correctly merged into the shading map~\cite{BigTimeLi18}. However, it cannot be extracted from the shading without shape information, so the results of decomposition cannot be re-rendered into a new image. Re-rendering using the decomposed intrinsic components can be achieved by choosing objects with simple shapes~\cite{LIME,sfsnet,SIRS}. In more recent research, attention has been put on decomposing shape, reflectance, and lighting on a complete scene~\cite{yu19inverserendernet,neuralSengupta19}, by still assuming Lambertian reflectance (represented by an albedo map). It becomes rather challenging to consider more realistic reflections such as specular highlights, shadows, and interreflection for single image intrinsic decomposition at the scene context, because the diverse objects and their complex interactions significantly increase the ambiguity of the solution space.  Recently, a complex self-supervised outdoor scene decomposition framework~\cite{yu2020selfrelight} has been proposed to further separate shadows from the shading image with a multi-view training dataset, and relighting a city scene has been achieved by using a neural renderer with learned shape and lighting descriptors~\cite{liu2020factorize}.

%% emphasize issues about dataset
Lacking an appropriate dataset is the main obstacle that prevents a deep neural network from learning intrinsic decomposition at the scene level effectively. Existing datasets are either sparsely annotated (\eg,~IIW~\cite{bell14intrinsic} and SAW~\cite{saw}) or partially photorealistic with noisy appearance (\eg,~SUNCG~\cite{suncg}, CGIntrinsics~\cite{li2018cgintrinsics}, PBRS~\cite{zhang2016physically}). We notice that the recently released FSVG dataset~\cite{gta5} created
from videogames\footnote{We want to comment that the computer graphics (CG) rendering and physics-based image formation (IF) are different and approximation is widely applied in graphics for efficiency consideration. Therefore, we mainly focus on the components which CG and IF render in the same way, such as diffuse shading and shadow. For components that CG does not follow IF (\eg,~specular component and interreflection), we will not discuss them in this work.} complements well existing intrinsic image decomposition datasets in several aspects: 1) it contains dense annotations of albedo and depth maps (from which shape information can be acquired); 2) the renderings are of high quality with complex reflection effects and little noise; and 3) it covers unprecedentedly abundant outdoor urban scenes on a city scale. However, it is non-trivial to directly apply this data to the task of intrinsic image decomposition. The dataset is initially collected for high-level vision tasks such as segmentation and recognition, therefore only the albedo is directly usable for supervising decomposing intrinsic components.

%% abstract and claim contributions
In this paper, we propose \textbf{DeRenderNet} to learn intrinsic image decomposition from the outdoor urban scenes based on FSVG dataset~\cite{gta5}. As the name suggests, we ``derender" the scene by \textbf{de}composing the scene intrinsic components to obtain albedo, lighting, and shape-independent shading (which mainly contains cast shadows), and then \textbf{render}ing shape-dependent shading (dot product of normal and lighting) using a self-supervised neural \textbf{net}work. Our major contributions include:
\begin{itemize}
   \item
     We propose an image formation model (Section \ref{sec:image_formation_model}) and data preprocessing pipeline (Section \ref{sec:training_data}) that takes full advantage of FSVG dataset~\cite{gta5} to conduct intrinsic image decomposition for outdoor urban scenes.
   \item
     We design a two-stage network (Section \ref{sec:proposed_mathod}) to effectively extract four types of intrinsic components, including albedo, lighting in its latent space, shape-independent shading, and re-rendered shape-dependent shading with the guidance of the depth map by self-supervised learning.  
   \item
     We demonstrate that our method avoids over-smoothing the albedo map by correctly decomposing shape-(in)dependent shadings (Section \ref{sec:eva}). The recovered shadow-free albedo with clean details could benefit high-level vision tasks for urban scenes (Section \ref{sec:application}).
\end{itemize}

%--------------------------------------------------------------------------------------------------------

\begin{figure*}
  \centering
    \includegraphics[width=2\columnwidth]{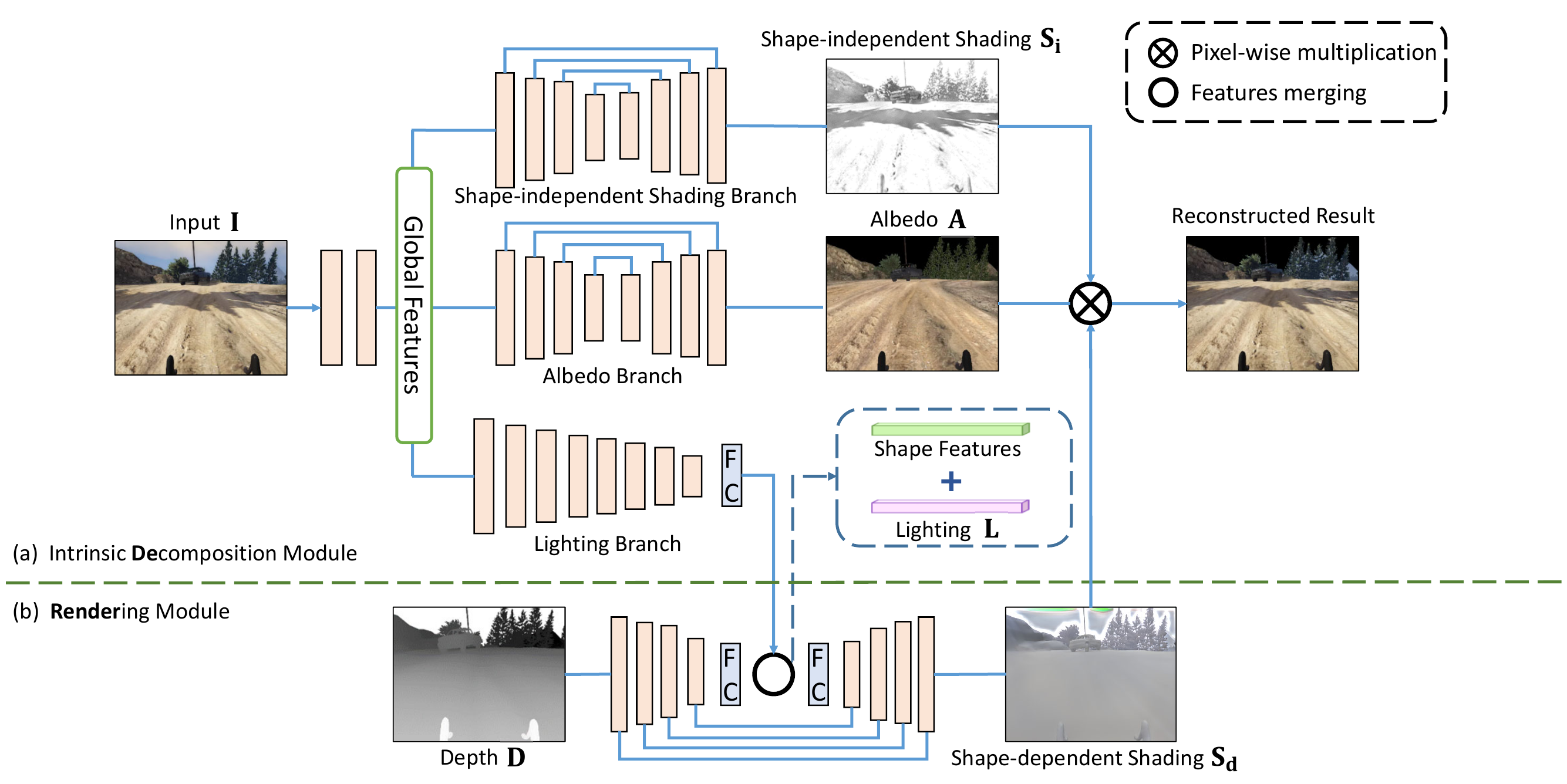}
    \caption{The network architecture of DeRenderNet is composed of two modules: (a) Intrinsic Decomposition Module and (b) Rendering Module. Kernel sizes for all convolutional layers are 5$\times$5. The intrinsic decomposition module takes an image $\bI$ as input and estimates the intrinsic components: shape-independent shading $\bS_{\rm{i}}$, albedo $\bA$, and latent lighting code $\bL$. The rendering module first takes the depth image $\bD$ as input and extracts shape features in latent space, and then it is concatenated with $\bL$ to decode the shape-dependent shading $\bS_{\rm{d}}$. The estimated $\bS_{\rm{i}}$, $\bA$, and $\bS_{\rm{d}}$ can be used to generate a reconstructed image via pixel-wise multiplication.}
    \label{fig:network}
\end{figure*}

%--------------------------------------------------------------------------------------------------------

\section{Related Work}
\label{sec:related_work}
Recently, learning-based methods have shown advantages over conventional methods relying on hand-crafted priors. In this section, we discuss current learning-based methods for intrinsic image decomposition and, more broadly, those for inverse rendering.

\subsection{Intrinsic Image Decomposition}
\label{subsec:intrinsic_image_decomposition}
The traditional intrinsic image decomposition refers to decompose an image into reflectance and shading.
However, even for such a two-layer decomposition, ground truth data covering diverse scenes are difficult to collect. There are only a few datasets for this problem, such as
single-object-based MIT intrinsic dataset~\cite{grosse2009ground}, animation-movie-based MPI-Sintel dataset~\cite{butler2012naturalistic},
sparse manually labeled IIW~\cite{bell14intrinsic} and SAW datasets~\cite{saw}, OpenGL-rendered SUNCG dataset~\cite{suncg}, and physically rendered CGIntrinsics dataset~\cite{li2018cgintrinsics}.
Among them, MIT, IIW, and SAW contain real captured images, MPI-Sintel, SUNCG, and CGIntrinsics are synthetic datasets. 
To date, the majority of intrinsic methods~\cite{narihira2015learning,zhou2015learning,2017reflectanceFiltering} used the IIW dataset~\cite{bell14intrinsic}  for training and validation. Narihira~\etal~\cite{narihira2015learning} extracted deep features from two patches and trained a classifier to determine the pairwise lightness ordering.  Zhou~\etal~\cite{zhou2015learning} proposed a data-driven method to predict lightness ordering and integrated it into energy functions. Nestmeyer~\etal~\cite{2017reflectanceFiltering} used signal processing techniques and a bilateral filter to get reasonable decomposition results. These methods rely on pairwise reflectance comparison to guide the network prediction due to the lack of densely labeled ground truth albedo. 
Li and Snavely~\cite{li2018cgintrinsics} proposed a deep model to combine CGIntrinsics, IIW, and SAW datasets to learn intrinsic decomposition with densely labeled synthetic data.
Later, they~\cite{BigTimeLi18} proposed an image sequence dataset with a fixed viewpoint to learn constant reflectance images over time and successfully put lighting color into shading images.
Bonneel~\etal~\cite{bonneel2017intrinsic} reviewed and evaluate some past intrinsic works for image editing.
Fan~\etal~\cite{fan2018revisiting} revisited this problem and used a guiding network to finetune the albedo estimation results. Liu~\etal~\cite{liu2020factorize} learned intrinsic decomposition for relighting city scenes on a time-lapse dataset collected from Google Street View.

In this paper, we also consider the colors in shading and further decompose shading into shape-dependent and shape-independent terms.

\subsection{Inverse Rendering}
\label{subsec:inverse_rendering}
The appearance of an observed image is jointly determined by shape, reflectance, and lighting. Previous works tried to take such components as input and train a neural network to estimate the result of forward rendering~\cite{nalbach2017deep}, or formulate it in a differentiable way~\cite{zhao2020physics}. Compared with forward rendering, inverse rendering is more widely used in computer vision applications, but it is also a difficult task. To alleviate the difficulty in joint intrinsic components estimation for general objects, single types of objects with relatively unified material property (such as human faces) have been studied. Sengupta~\etal~\cite{sfsnet} assumed the human face is a Lambertian object and used supervised training on synthetic data, while later used real images for finetuning using self-supervised learning. Yamaguchi~\etal~\cite{yamaguchi2018high} proposed a hybrid reflectance model that combined Lambertian and specular properties to get a more realistic face appearance. Meka~\etal~\cite{LIME} used an encoder-decoder architecture to successfully estimate the material of single objects. Recently, research focus has turned to inverse rendering of a scene. For example, Yu~\etal~\cite{yu19inverserendernet} used multi-view stereo (MVS) to get a rough geometry and then estimated a dense normal map by assuming a Lambertian model and albedo priors on outdoor scenes. Yu~\etal~\cite{yu2020selfrelight} further estimated a global shadow map to get shadow-free albedo in order to relight outdoor scenes. Sengupta~\etal~\cite{neuralSengupta19} rendered indoor scenes with the Phong model and proposed a learning-based approach to jointly estimate albedo, normal, and lighting of an indoor image based on a synthetic dataset. However, reflectance on the scene level is much more complex than what a Lambertian or Phong model can describe. Li~\etal~\cite{li2020inverse} proposed a physically-based renderer with SVBRDF to render more realistic images to learn inverse rendering for complex indoor scenes.

In this paper, we extract intrinsic components to inverse-render a scene without explicitly estimating depth and specularity; instead we focus on extracting an independent shadow layer.

%--------------------------------------------------------------------------------------------------------

\section{Image Formation Model}
\label{sec:image_formation_model}
The goal of DeRenderNet is to decompose an image in its intrinsic components for real-world outdoor scenes, containing complex shape, reflectance, and lighting (shadows). To improve the classic formulation which uses the albedo-scaled shading, we define the \emph{shape-dependent shading} that is consistent with classic shading as the dot product between normal and lighting, and the \emph{shape-independent shading}, which mainly contains cast shadows (which are not determined by the shape of an object itself but by the occlusion on the light path). We ignore the specular highlights and other global illumination effects (such as inter-reflection, transparency and translucency), since they are too complex in a scene with many objects. Fortunately, they are generally sparse, so we leave them as additive noise. Then the image formation can be expressed as
\begin{equation}
\bI = \bA \odot \bS_\bd(\bD,\bL) \odot \bS_\bi + \epsilon,
\label{equ:Formation}
\end{equation}
where albedo map $\bA$ and shape-dependent shading $\bS_\bd$ have the same physical meaning as other intrinsic decomposition works~\cite{li2018cgintrinsics, yu19inverserendernet}. To link our definition to diffuse shading, which is defined as $\tilde \bS_\bd(\bn, \Omega)=\Sigma_k \max(\bn^\top \omega_k, 0)$ ($\omega_k \in \Omega$ is a 3-D vector encoding directional light intensity and direction sampled from the visible hemisphere of an environment map $\Omega$), we design a rendering module $\Theta$ that takes depth $\bD$ and lighting $\bL$ in a latent space as input to render $\bS_\bd$ instead of writing down its explicit form (usually not differentiable), \ie, $\tilde \bS_\bd(\bn, \Omega) =  \tilde \bS_\bd(\Theta(\bD, \bL)) = \bS_\bd(\bD, \bL)$. $\bS_\bd$ does not take any global lighting effect such as shadows or interreflections into account, while shape-independent shading $\bS_\bi$, as the complement of $\bS_\bd$, is intended to encode the contribution of cast shadows and interreflections.

Given a scene image $\bI$ as input, we decompose all intrinsic components $\bA$, $\bS_{\bd}$, $\bS_{\bi}$, with $\bD$ and $\bL$ as latent variables in~\Eref{equ:Formation}, as illustrated in~\Fref{fig:network}. We do not explicitly estimate the shape as one of the intrinsic components because inferring the scene depth from a single image is another highly ill-posed problem~\cite{MDLi18,qi2018geonet}. Due to the strong shape-light ambiguity in scenes, estimating the shape information constrained by reconstruction errors~\cite{neuralSengupta19,yu19inverserendernet} is fragile. 

%--------------------------------------------------------------------------------------------------------

\section{Training Data}
\label{sec:training_data}
To achieve shape-(in)dependent decomposition on shading, we need densely labeled depth maps in addition to albedo. We therefore propose a data preprocessing method to obtain additional supervision for our task.

\subsection{FSVG Dataset}
The FSVG dataset~\cite{gta5} is created by collecting and labeling an enormous amount of data (about 220k images) from videogames such as GTAV and The Witcher 3. It includes instance and semantic segmentation labels, densely labeled depth, optical flow, and pixel-wise albedo values. This dataset is useful in many high-level vision tasks thanks to both realistically rendered images and accurate ground truth labels. By collecting data from videogames, a large number of images in outdoor scenes with great diversity can be acquired, which is highly difficult and impractical using human labor. 
In this paper, we use densely labeled ground truth values of albedo and depth for directly and indirectly supervising intrinsic image decomposition task.
% In addition, the FSVG dataset~\cite{gta5} also provides densely labeled ground truth values for albedo and depth, which are useful in directly or indirectly supervising intrinsic image decomposition. 

\subsection{Data Preprocessing}
\begin{figure}
  \centering
    \includegraphics[width=1\columnwidth]{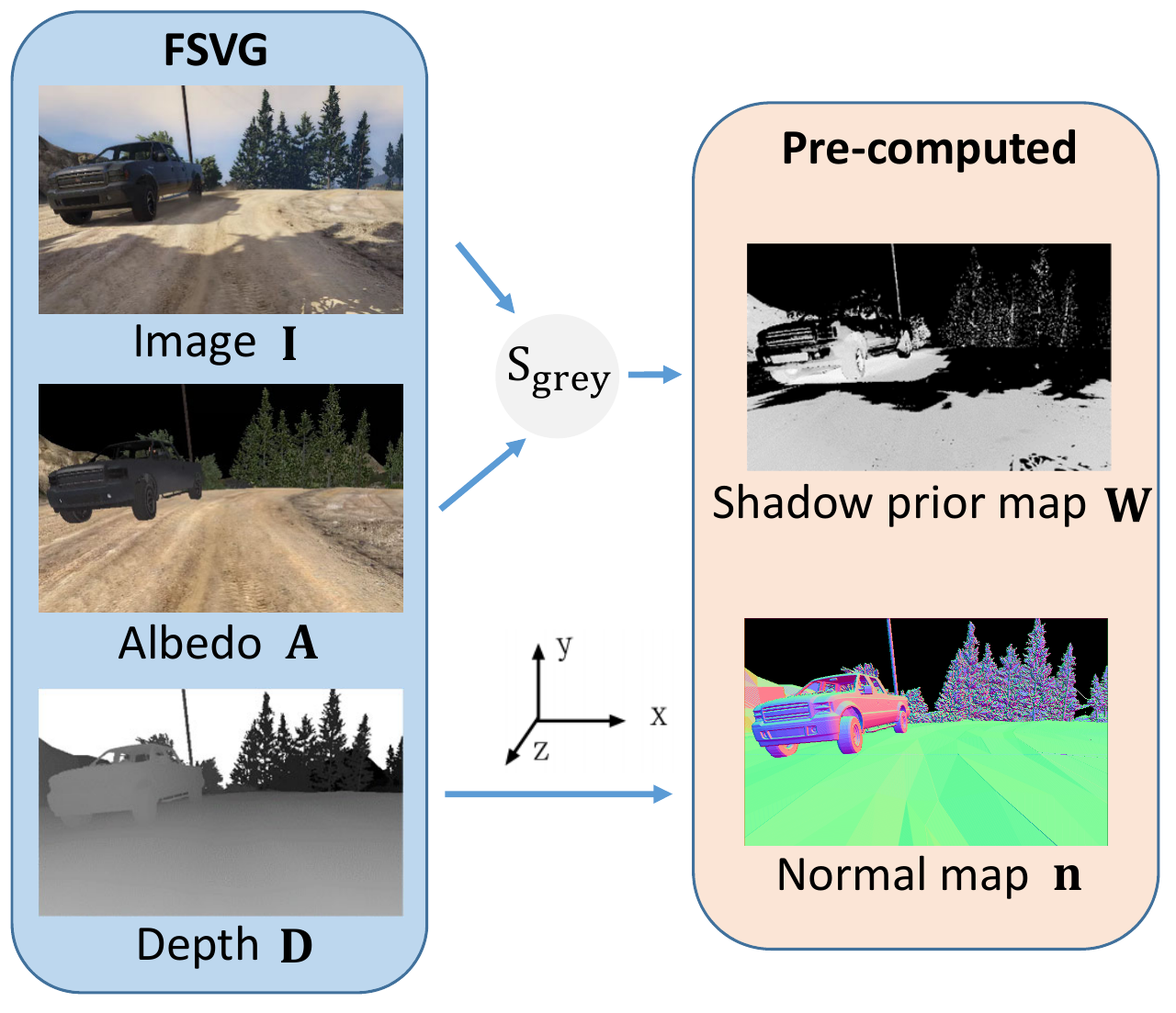}
    \caption{Our training data contains two parts: the image, albedo, and depth directly available from FSVG dataset~\cite{gta5} (left column), and the shadow prior map and normal map (right column) pre-computed from the data on the left.}
    \label{fig:data}
\end{figure}

To provide more useful supervision for our task, we pre-compute two types of additional labels: a shadow prior map and a normal map, as shown in~\Fref{fig:data}. 

The shadow prior map provides an approximate probability distribution of shadow regions. To obtain the shadow prior map, we first calculate the pseudo-shading as $\bS_{\rm{pseudo}} = {\bI}/{\bA} $. Then we set the regions where $\bS_{\rm{pseudo}} > 1$ to a large intensity value ($10$ in our experiment) to eliminate the effects of non-shadow parts and convert $\bS_{\rm{pseudo}}$ into a greyscale image $\bS_{\rm{grey}}$.
We define the shadow prior map as
\begin{equation}
\bW = \frac{1}{\sqrt{2\pi}}\exp(-\frac{\bS_{\rm{grey}}^2}{2}).
\end{equation}
The shadow prior map alone is not sufficient to distinguish dark shadings and shadows, so we consider to involve depth for further decomposition. The depth map contains pixel-wise relative distance information, but it does not explicitly represent surface orientations, which encode important cues for intrinsic image decomposition. We therefore calculate the pixel-wise normal vector according to the depth map. We assume a perspective camera model, so the projection from 3D coordinates $\mathbf{P} = (x,y,z)$ to 2D image coordinates $\mathbf{p} = (u,v,1)$ is given by
\begin{equation}
\mathbf{p} = \frac{1}{z}\mathbf{KP},\quad \mathbf{K} = \begin{bmatrix} f & 0 & c_x  \\ 0 & f & c_y \\ 0& 0& 1 \end{bmatrix}.
\end{equation}
Accordingly the surface normal directions can be calculated as $\bar{\bn}^\top = [-z_{x}(u,v),-z_{y}(u,v),1]$. We use unit-length normal vectors $\bn = \bar{\bn}/\|\bar{\bn}\|$.   

%--------------------------------------------------------------------------------------------------------

%--------------------------------------------------------------------------------------------------------

\section{Proposed Method}
\label{sec:proposed_mathod}
In this section, we show how to solve the intrinsic image decomposition defined in~\Eref{equ:Formation} by fully exploiting the available and pre-computed data, instead of relying on smoothness terms on albedo and shading~\cite{bell14intrinsic,li2018cgintrinsics,fan2018revisiting,2017reflectanceFiltering}. 

\subsection{Network Structure Design}
We have the following considerations: 1) we want to estimate as many as possible intrinsic components according to the data available; 2) we want to use depth to guide our intrinsic image decomposition but avoid using it as a necessary input; 3) we want to use learned shape features and lighting in latent space to overcome the limited representation power of parametric image formation models.

Based on the considerations above, we create a two-stage framework called \textbf{DeRenderNet}. The first stage of DeRenderNet is an intrinsic $\textbf{de}$composition module that takes an observed image $\bI$ as input and estimates three intrinsic components: albedo $\bA$, lighting $\bL$ in latent space, and shape-independent shading $\bS_\bi$. The second stage of DeRenderNet is a $\textbf{render}$ing module, which generates shape-dependent shading $\bS_\bd$ based on lighting code $\bL$ extracted by the previous module and depth map $\bD$. 

\subsubsection{Intrinsic Decomposition Module}
The intrinsic decomposition module is a single-input-multi-output network that consists of three branches, as shown in~\Fref{fig:network} (a). The first two layers of the network are used to extract global features from an input image $\bI$, and then two residual block based architectures are used to predict $\bA$ and $\bS_\bi$, respectively. The skip connections in residual blocks allow the high-frequency information to be merged with low-frequency features. Since $\bL$ is a latent code, we use a deep encoder followed by a fully connected layer to predict it. We find that such an architecture can help separating $\bS_\bi$ and $\bS_\bd$, because $\bS_\bi$ has stronger gradients on edges. These features can pass through convolutional layers but are not preserved after traversing the fully connected layer. In this way, the intrinsic decomposition module can partly separate $\bS_\bi$ and $\bS_\bd$ features. 

\subsubsection{Rendering Module}
The rendering module is a multi-input-single-output network. It takes depth $\bD$ and lighting code $\bL$ as input, and outputs $\bS_\bd$. We treat this network as a ``rendering engine" that is able to generate realistic shading for complex scenes. We design it to handle shape-related complex reflection in a self-learning way. The main structure is an encoder-decoder, as shown in~\Fref{fig:network} (b). The encoder part is used to encode a shape prior and to extract features of shape information. Shape features obtained by the last layer of the encoder are concatenated with lighting code $\bL$ before being sent into the decoder part to generate $\bS_\bd$. We use a fully connected layer at the end of the encoder and the beginning of the decoder to extract high-frequency features and merge these two types of features together. Several skip connections are used to make full use of shape information.

\subsection{Learning Shape-(in)dependent Shadings}
DeRenderNet is trained with the labeled and pre-computed data in~\Fref{fig:data}. We use ground truth albedo labels and dense depth labels from the FSVG dataset~\cite{gta5} to supervise the albedo estimation and guide the self-supervised learning, respectively. 
Since the ground truth lighting, shading, and shadows are not available in current datasets, we propose a self-supervised learning method to estimate lighting code $\bL$, shape-dependent shading $\bS_\bd$, and shape-independent shading $\bS_\bi$ by designing appropriate loss functions.

\subsubsection{Supervised Loss}
Since the images in the FSVG dataset~\cite{gta5} are equipped with ground truth albedo values, we use the albedo as direct supervision to train our intrinsic decomposition module:
\begin{equation}
  \mathcal{L}_{\rm{a}} = \frac{1}{M}\sum_{i=1}^{M}(\bA_i - \bA_i^{\ast})^2 ,
  \label{equ:supervised}
\end{equation}
where $\bA$ is the estimation of albedo image, $\bA^{\ast}$ is the ground truth value
of albedo, $M$ is the total number of valid pixels correspond to non-sky regions.

\subsubsection{Self-supervised Losses}
Although $\bA^{\ast}$ provides direct supervision for albedo estimation, there is no supervision guidance for the other two intrinsic attributes: $\bL$ and $\bS_\bi$. Hence we propose a self-supervised method to make our intrinsic decomposition module learn how to effectively extract these two components and to make our rendering module learn how to render $\bS_\bd$ based on the estimated $\bL$ and the guiding depth map $\bD$ at the same time. 

\smallskip
\noindent\textbf{Shape-independent Shading Loss:}
We use the pre-computed shadow prior map, which is an approximate probability distribution indicating where shadows should appear, to guide the prediction of shape-independent shading $\bS_i$ using the following loss function:
\begin{equation}
\mathcal{L}_{\rm{si}} = \frac{1}{M}\sum_{i=1}^{M}w_i\sum_{j\in \mathcal{N}(i)}({\bS_\bi}_i - {\bS_\bi}_j)^2.
\label{equ:shadow}
\end{equation}
Here $w_i \in \bW$ is the shadow prior at pixel $i$, $\mathcal{N}(i)$ denotes the neighborhood of the pixel at position $i$, and $M$ is the same as in~\Eref{equ:supervised}. This loss function mainly encourages shape-independent shading $\bS_\bi$ to be considered as shadows, and we will further decouple it from the dependency on shape by the next loss function. 

\smallskip
\noindent\textbf{Shape-dependent Shading Loss:}
After putting high-frequency terms as an additive noise in $\epsilon$, it is natural to assume that the shading is smooth. More specifically, considering that the outdoor light source mainly comes from the distant sun, albedo-normalized scene radiance values on the same shadow-free plane should be roughly consistent. On the other hand, though using $\bD$ and $\bL$ significantly suppresses the edges of shadows, there still exists ambiguity in non-edge parts. To further remove this ambiguity, we define the shape-dependent shading loss as
\begin{align}
  \mathcal{L}_{\rm{sd}} = ~&\frac{1}{M}\sum_{i=1}^{M}[({\bS_\bd}_i(\bD,\bL) - \sum_{m=1}^{B}{\bn_i}^\top\bl_{m})^2 +  \nonumber\\
                  ~&(1-w_i)(\bA_i{\bS_\bd}_i(\bD,\bL)-\bI_i)^2],
\label{equ:shading}
\end{align}
where $\bn_i$ is the normal vector at pixel $i$ of image, $\bl_m$ is a directional light (a $3$-D vector) extracted from the latent code $\bL$ (in total $B$ vectors are extracted) to encourage similar surface normals to have similar ``shading'' under $\bL$, $w_i$ is the same as in~\Eref{equ:shadow}, and $M$ is the same as in~\Eref{equ:supervised}. In particular, we split $\bL$ as a total of $B$ $4$-D vectors and take the first three elements of each $4$-D vector as $3$-D vector $\bl_{m}$. The first term in this loss aims to smooth the shape-dependent shading approximately in a planar surface, and the second term is used to suppress shadows in the rendering process of shape-dependent shadings. 

\smallskip
\noindent\textbf{Reconstruction Loss:}
Given a estimated albedo $\bA$, a shape-dependent shading $\bS_\bd$ and a shape-independent shading $\bS_\bi$, we compute a reconstructed image using \Eref{equ:Formation}. Then we use a pixel-wise $L^2$ loss between the reconstructed image and the input image as the reconstruction loss:
\begin{equation}
\mathcal{L}_{\rm{recon}} = \frac{1}{M}\sum_{i=1}^{M}(\bA_i{\bS_\bd}_i(\bD,\bL){\bS_\bi}_i - \bI_i)^2.
\label{equ:recon}
\end{equation}

\subsection{Implementation Details}
We train our DeRenderNet to minimize
\begin{equation}
\mathcal{L} = \mathcal{L}_{\rm{recon}} + \lambda_1 \mathcal{L}_{\rm{a}} + \lambda_2 \mathcal{L}_{\rm{sd}} + \lambda_3 \mathcal{L}_{\rm{si}},
\label{equ:full}
\end{equation}
where $\lambda_1 = 0.8$ and ~$\lambda_2, ~\lambda_3 = 0.5$. We use all 220k training images from the FSVG dataset~\cite{gta5} to train our networks. Before training, the images are resized to $320\times240$ due to the fully connected layers in DeRenderNet. For the middle layers, we use ReLU and Batch Normalization followed by convolutional layers. We use the same learning rate for all layers. According to previous lighting estimation studies~\cite{weber2018learning,hold2019deep}, the scene lighting information encoded in a latent space has a strong representation power, so we adopt the similar strategy. To figure out a proper number of dimensions of lighting code, experiments are conducted when the number of dimensions are searched from 0 to 2000. We empirically set the dimension of the latent lighting code to 1024 and $B$ in \Eref{equ:shading} as 256 based on the results from validation set. 

Please note that we only use depth as input in the training stage to guide DeRenderNet to learn how to extract $\bS_\bd$ and $\bS_\bi$ from images. To analyze the influence of depth errors during training, we added some noise to the depth input. However, we found that this will not affect our results, probably because the useful features could be extracted by the network despite noisy input depth. Once the training is completed, we can use the intrinsic decomposition module to get $\bS_\bi$ and $\bA$ without depth input. Please refer to our supplementary material to see the shape-dependent shading results with the estimated depth from MegaDepth~\cite{MDLi18} trained on in-the-wild data.

%--------------------------------------------------------------------------------------------------------

\begin{figure*}
    \centering
        \begin{tabular}{ccccccc}
            \hspace{-0\columnwidth}\includegraphics[width=0.323\columnwidth]{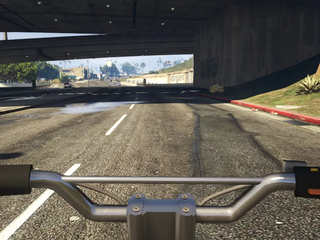} &
            \hspace{-0.04\columnwidth}\includegraphics[width=0.323\columnwidth]{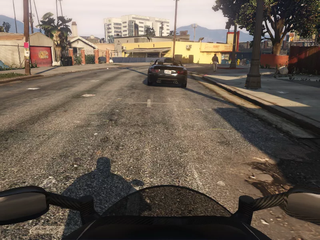} &
            \hspace{-0.04\columnwidth}\includegraphics[width=0.323\columnwidth]{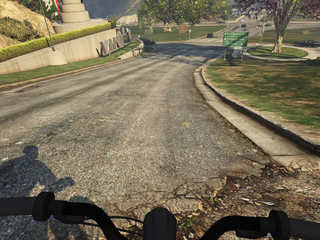} &
            \hspace{-0.04\columnwidth}\includegraphics[width=0.323\columnwidth]{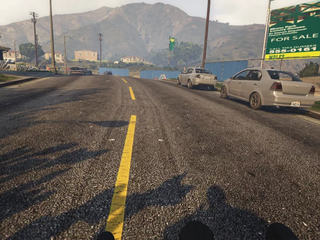} &
            \hspace{-0.04\columnwidth}\includegraphics[width=0.323\columnwidth]{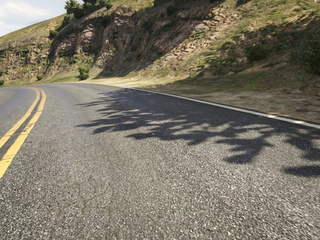} &
            \hspace{-0.04\columnwidth}\includegraphics[width=0.323\columnwidth]{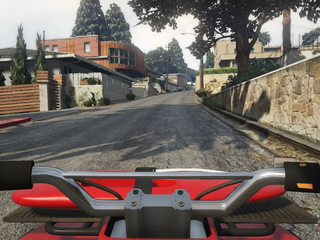} &
            \hspace{-0.03\columnwidth}\raisebox{0.16\columnwidth}{\rotatebox{270}{{\scriptsize Input}}} \\
            
            \hspace{-0\columnwidth}\includegraphics[width=0.323\columnwidth]{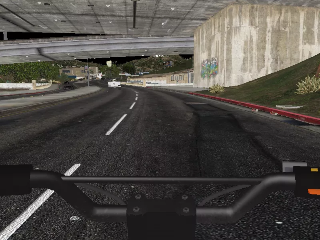} &
            \hspace{-0.04\columnwidth}\includegraphics[width=0.323\columnwidth]{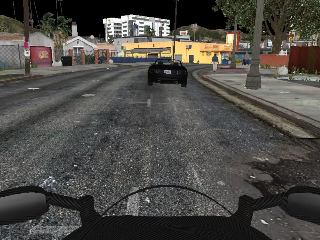} &
            \hspace{-0.04\columnwidth}\includegraphics[width=0.323\columnwidth]{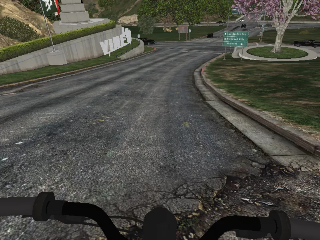} &
            \hspace{-0.04\columnwidth}\includegraphics[width=0.323\columnwidth]{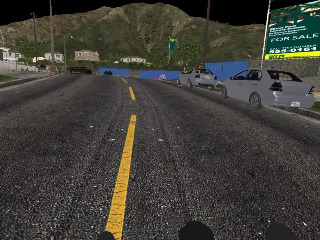} &
            \hspace{-0.04\columnwidth}\includegraphics[width=0.323\columnwidth]{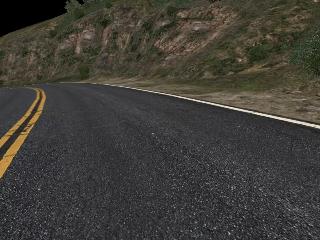} &
            \hspace{-0.04\columnwidth}\includegraphics[width=0.323\columnwidth]{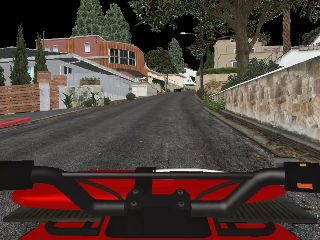} &
            \hspace{-0.03\columnwidth}\raisebox{0.165\columnwidth}{\rotatebox{270}{{\scriptsize GT ($\bA$)}}} \\
        
            \hspace{-0\columnwidth}\includegraphics[width=0.323\columnwidth]{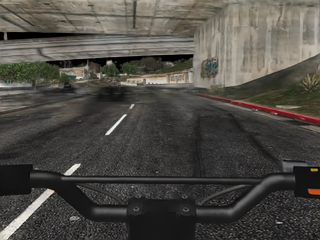} &
            \hspace{-0.04\columnwidth}\includegraphics[width=0.323\columnwidth]{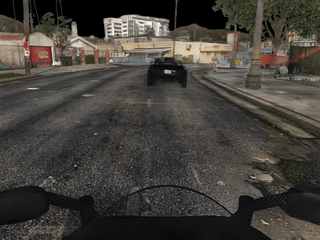} &
            \hspace{-0.04\columnwidth}\includegraphics[width=0.323\columnwidth]{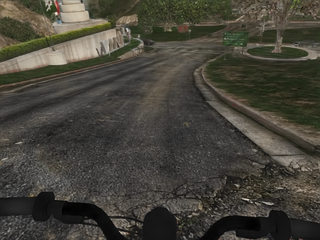} &
            \hspace{-0.04\columnwidth}\includegraphics[width=0.323\columnwidth]{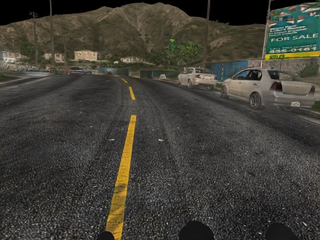} &
            \hspace{-0.04\columnwidth}\includegraphics[width=0.323\columnwidth]{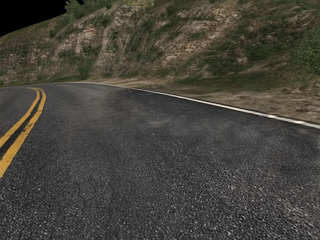} &
            \hspace{-0.04\columnwidth}\includegraphics[width=0.323\columnwidth]{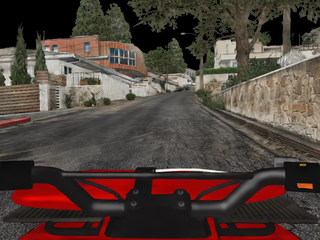} &
            \hspace{-0.03\columnwidth}\raisebox{0.185\columnwidth}{\rotatebox{270}{{\scriptsize Ours ($\bA$)}}} \\
            
            \hspace{-0\columnwidth}\includegraphics[width=0.323\columnwidth]{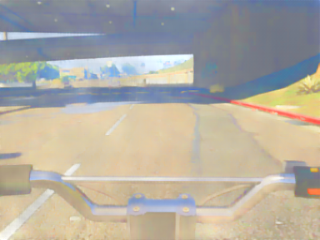} &
            \hspace{-0.04\columnwidth}\includegraphics[width=0.323\columnwidth]{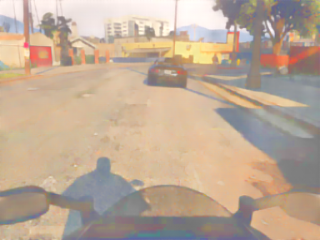} &
            \hspace{-0.04\columnwidth}\includegraphics[width=0.323\columnwidth]{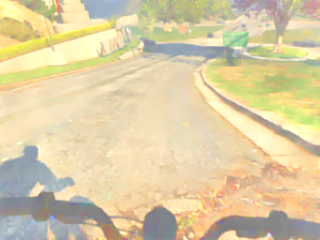} &
            \hspace{-0.04\columnwidth}\includegraphics[width=0.323\columnwidth]{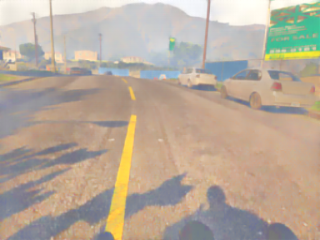} &
            \hspace{-0.04\columnwidth}\includegraphics[width=0.323\columnwidth]{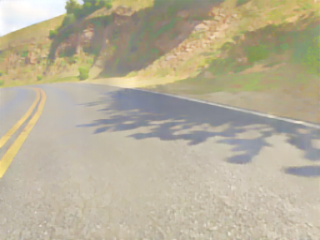} &
            \hspace{-0.04\columnwidth}\includegraphics[width=0.323\columnwidth]{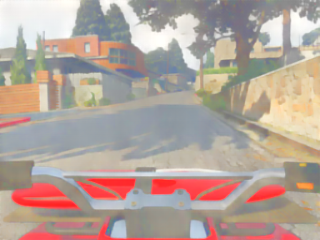} &
            \hspace{-0.03\columnwidth}\raisebox{0.21\columnwidth}{\rotatebox{270}{{\scriptsize Li~\etal~\cite{li2018cgintrinsics} ($\bA$)}}} \\
            
            \hspace{-0\columnwidth}\includegraphics[width=0.323\columnwidth]{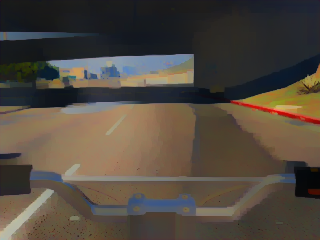} &
            \hspace{-0.04\columnwidth}\includegraphics[width=0.323\columnwidth]{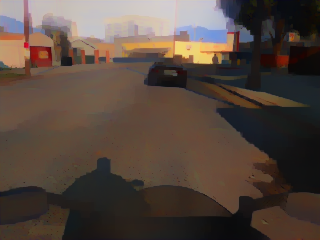} &
            \hspace{-0.04\columnwidth}\includegraphics[width=0.323\columnwidth]{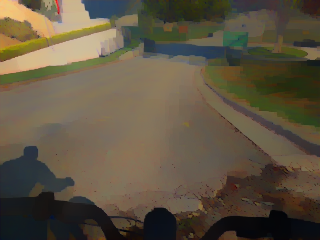} &
            \hspace{-0.04\columnwidth}\includegraphics[width=0.323\columnwidth]{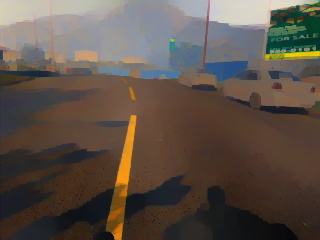} &
            \hspace{-0.04\columnwidth}\includegraphics[width=0.323\columnwidth]{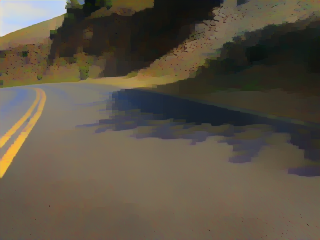} &
            \hspace{-0.04\columnwidth}\includegraphics[width=0.323\columnwidth]{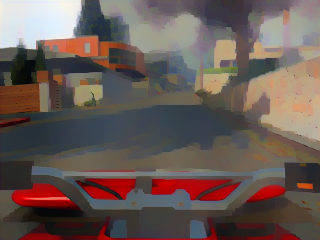} &
            \hspace{-0.03\columnwidth}\raisebox{0.22\columnwidth}{\rotatebox{270}{{\scriptsize Yu~\etal~\cite{yu19inverserendernet} ($\bA$)}}} \\
            
            \hspace{-0\columnwidth}\includegraphics[width=0.323\columnwidth]{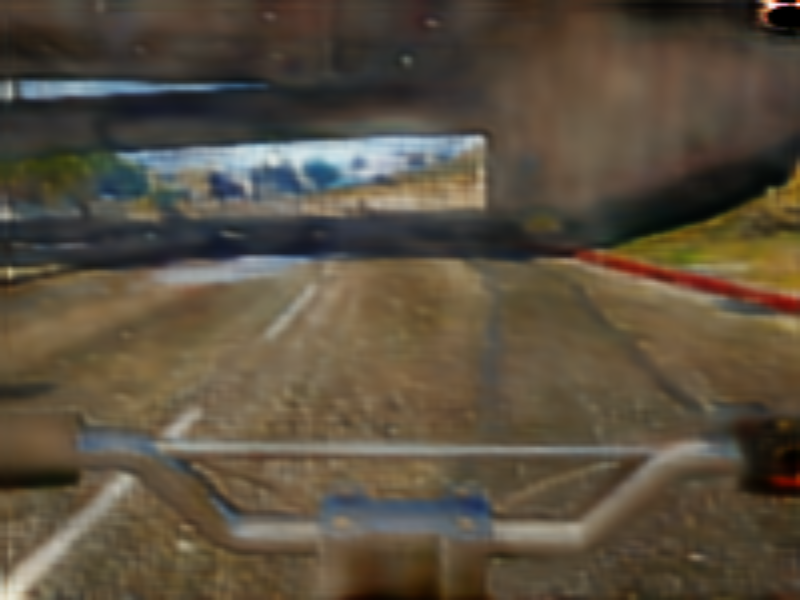} &
            \hspace{-0.04\columnwidth}\includegraphics[width=0.323\columnwidth]{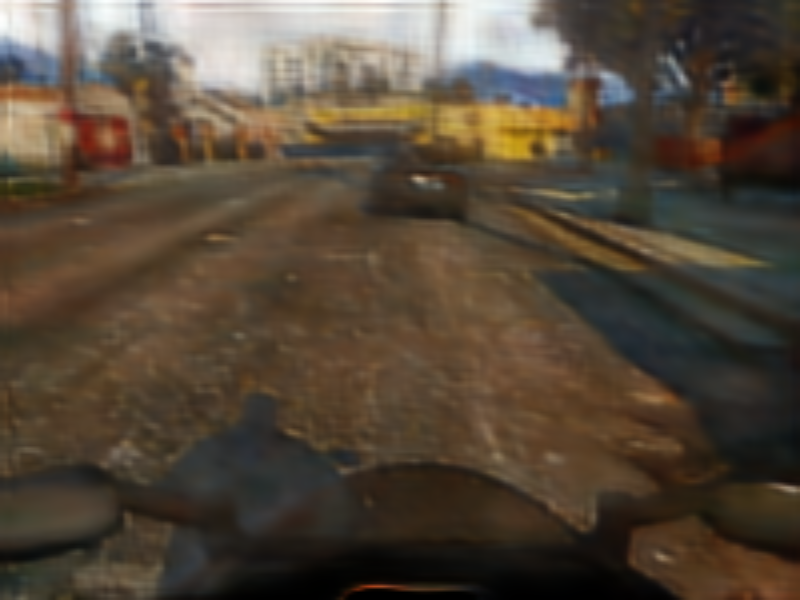} &
            \hspace{-0.04\columnwidth}\includegraphics[width=0.323\columnwidth]{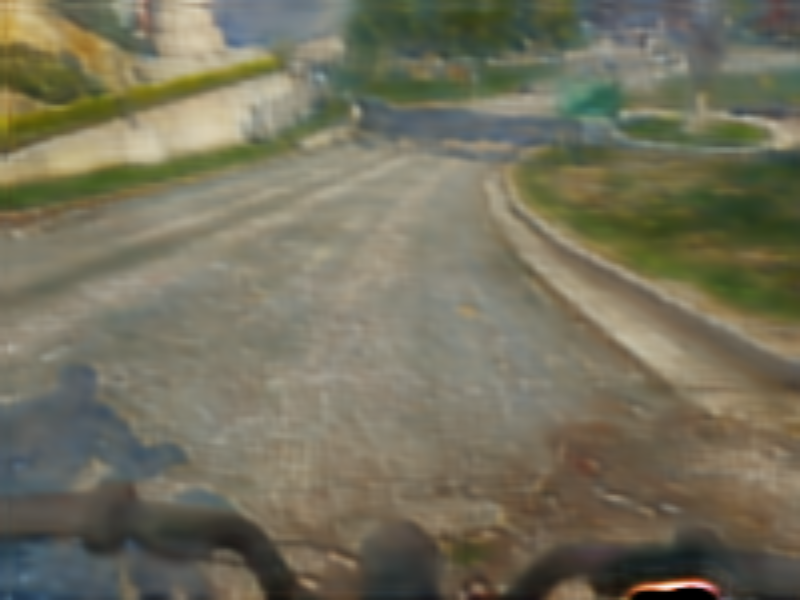} &
            \hspace{-0.04\columnwidth}\includegraphics[width=0.323\columnwidth]{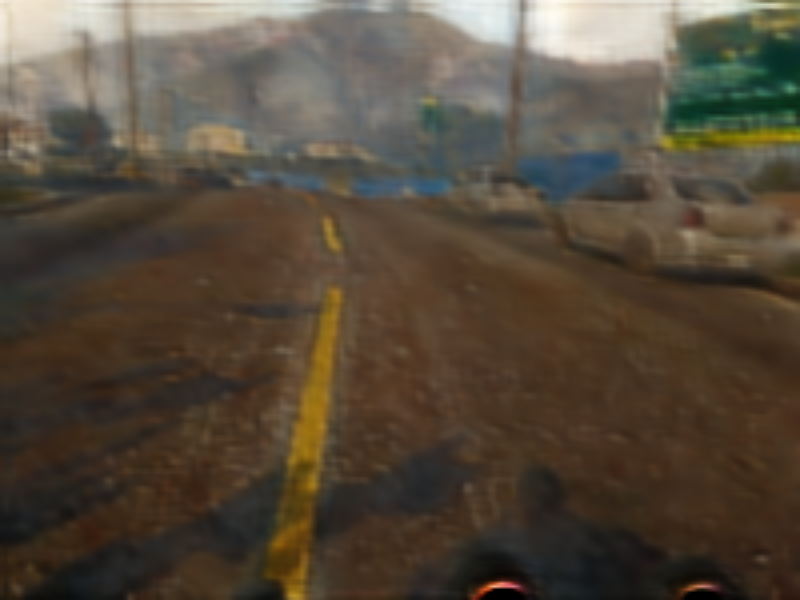} &
            \hspace{-0.04\columnwidth}\includegraphics[width=0.323\columnwidth]{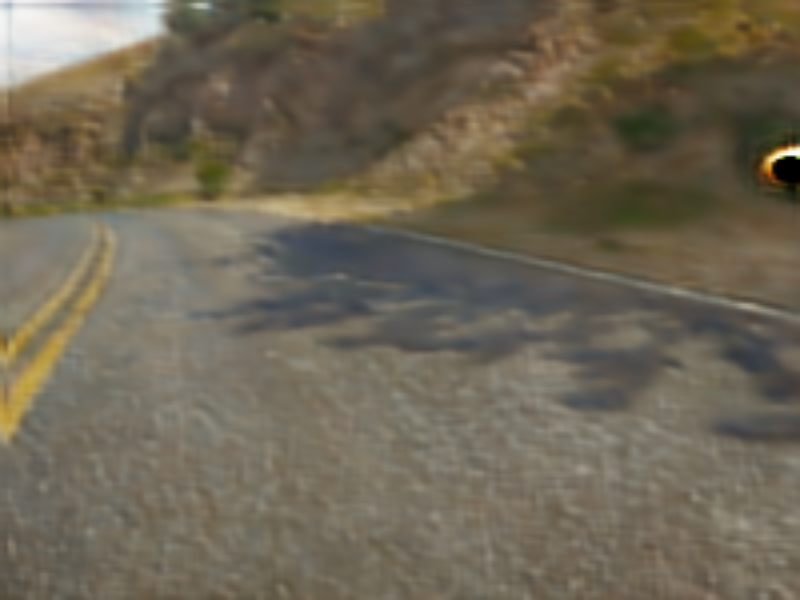} &
            \hspace{-0.04\columnwidth}\includegraphics[width=0.323\columnwidth]{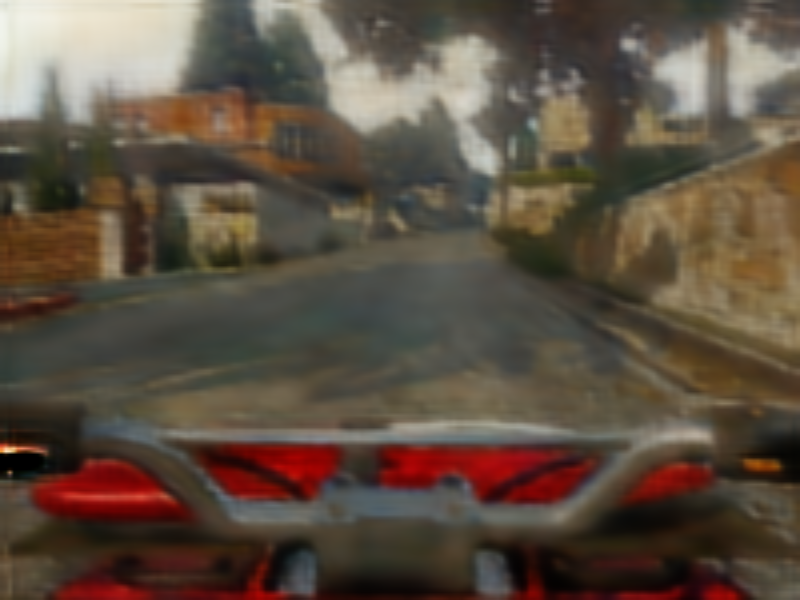} &
            \hspace{-0.03\columnwidth}\raisebox{0.22\columnwidth}{\rotatebox{270}{{\scriptsize Yu~\etal~\cite{yu2020selfrelight} ($\bA$)}}} \\
            
            \hspace{-0\columnwidth}\includegraphics[width=0.323\columnwidth]{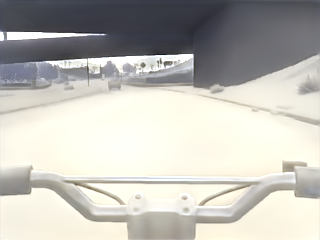} &
            \hspace{-0.04\columnwidth}\includegraphics[width=0.323\columnwidth]{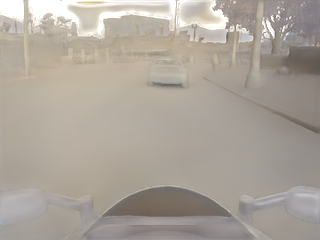} &
            \hspace{-0.04\columnwidth}\includegraphics[width=0.323\columnwidth]{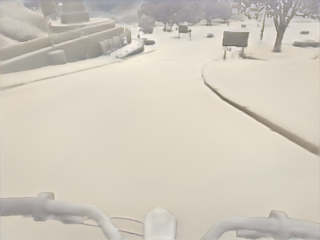} &
            \hspace{-0.04\columnwidth}\includegraphics[width=0.323\columnwidth]{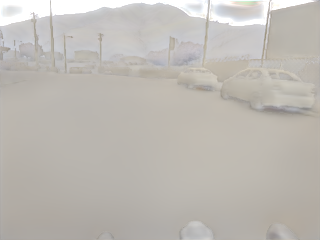} &
            \hspace{-0.04\columnwidth}\includegraphics[width=0.323\columnwidth]{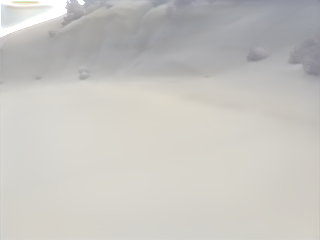} &
            \hspace{-0.04\columnwidth}\includegraphics[width=0.323\columnwidth]{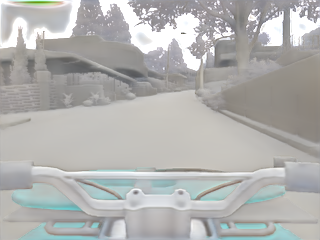} &
            \hspace{-0.03\columnwidth}\raisebox{0.185\columnwidth}{\rotatebox{270}{{\scriptsize Ours ($\bS_\bd$)}}} \\
            
            \hspace{-0\columnwidth}\includegraphics[width=0.323\columnwidth]{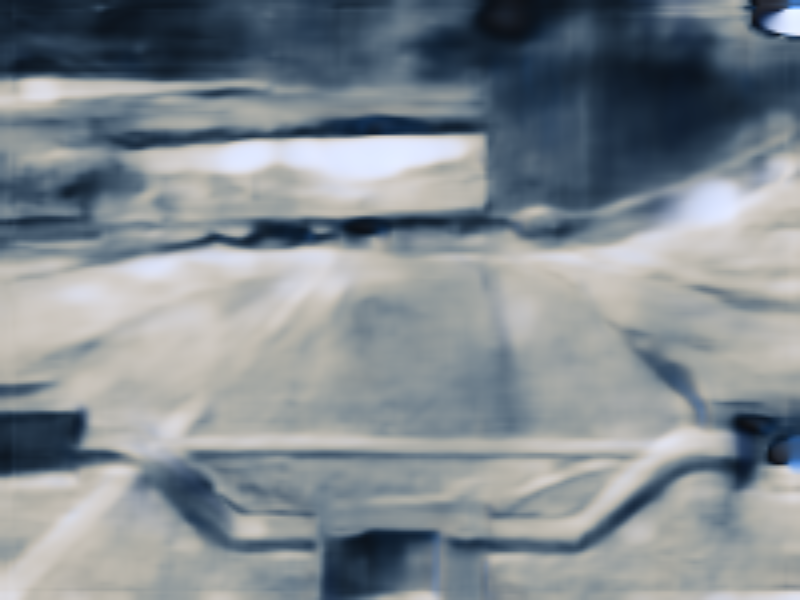} &
            \hspace{-0.04\columnwidth}\includegraphics[width=0.323\columnwidth]{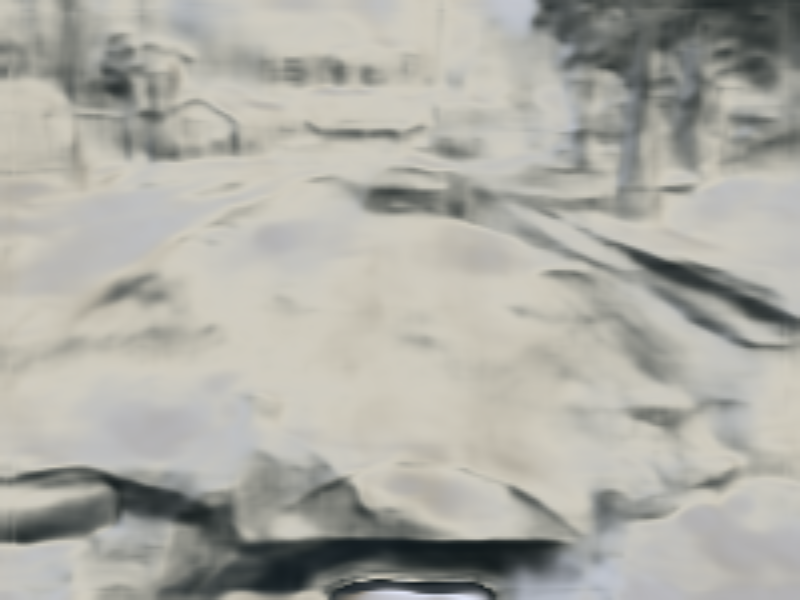} &
            \hspace{-0.04\columnwidth}\includegraphics[width=0.323\columnwidth]{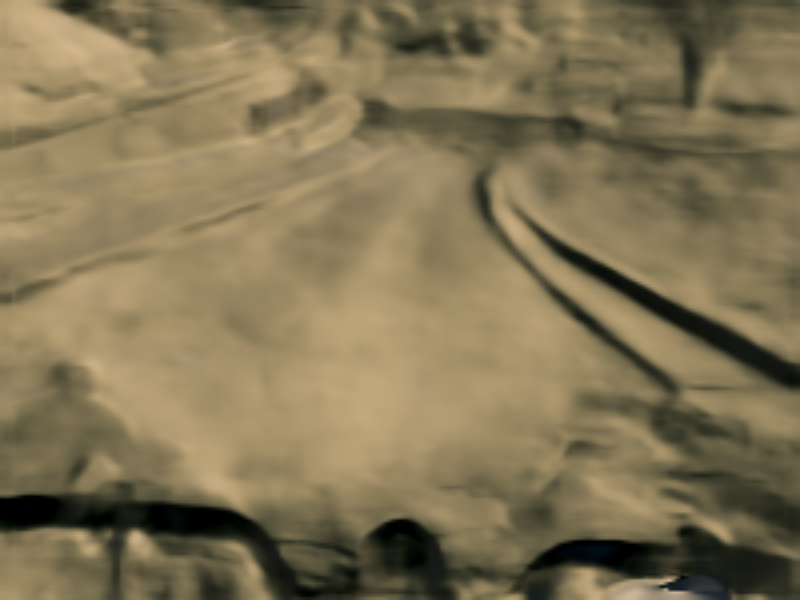} &
            \hspace{-0.04\columnwidth}\includegraphics[width=0.323\columnwidth]{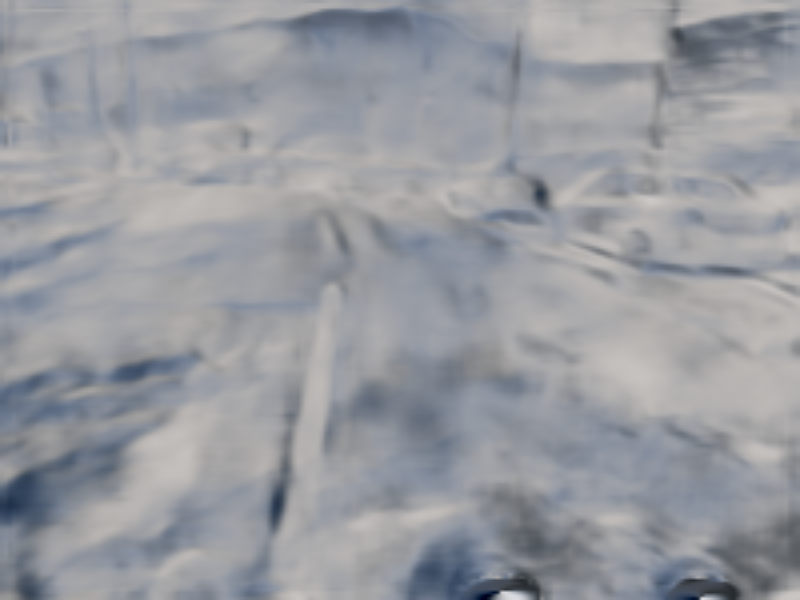} &
            \hspace{-0.04\columnwidth}\includegraphics[width=0.323\columnwidth]{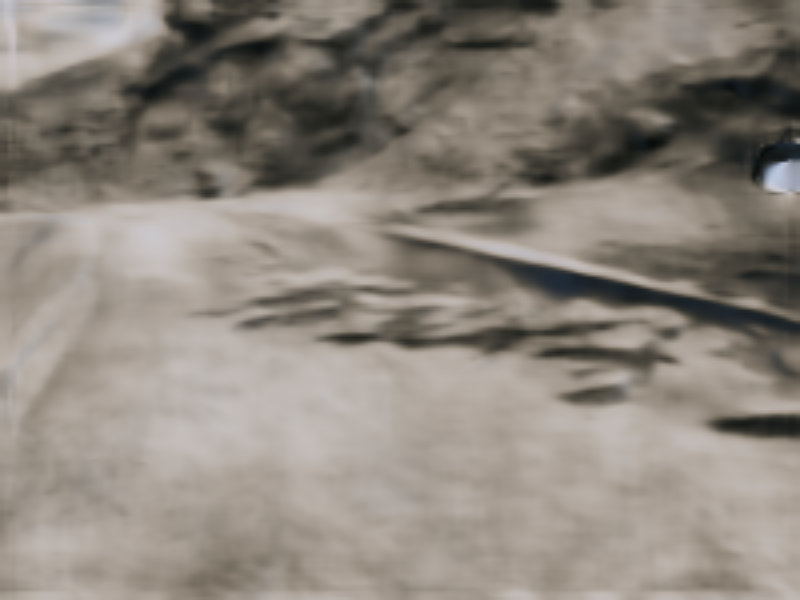} &
            \hspace{-0.04\columnwidth}\includegraphics[width=0.323\columnwidth]{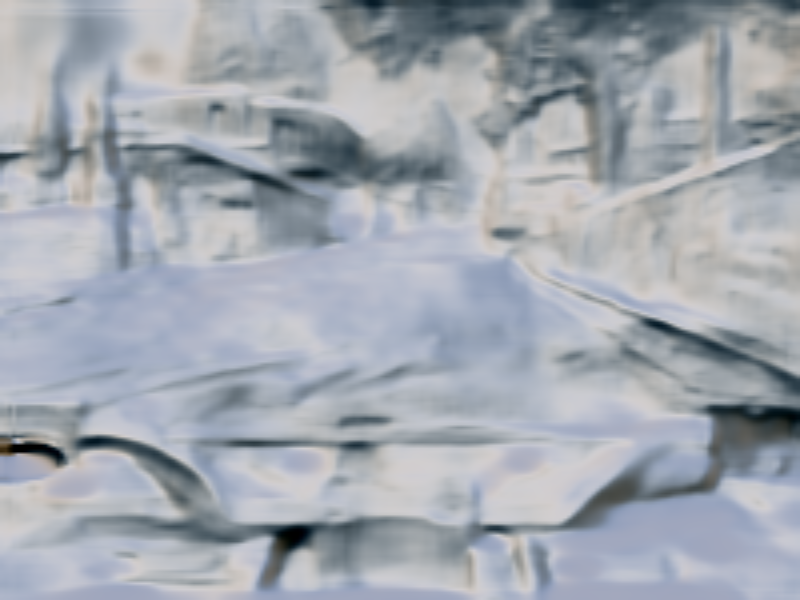} &
            \hspace{-0.03\columnwidth}\raisebox{0.22\columnwidth}{\rotatebox{270}{{\scriptsize Yu~\etal~\cite{yu2020selfrelight} ($\bS_\bd$)}}} \\
            
            \hspace{-0\columnwidth}\includegraphics[width=0.323\columnwidth]{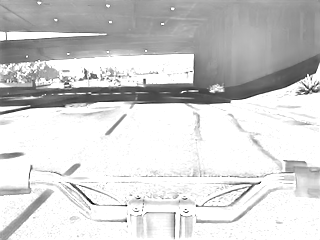} &
            \hspace{-0.04\columnwidth}\includegraphics[width=0.323\columnwidth]{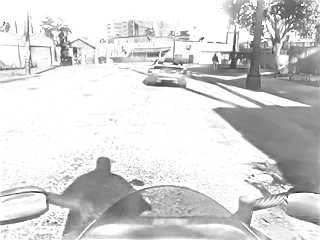} &
            \hspace{-0.04\columnwidth}\includegraphics[width=0.323\columnwidth]{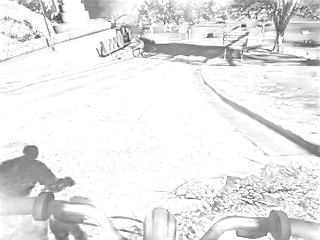} &
            \hspace{-0.04\columnwidth}\includegraphics[width=0.323\columnwidth]{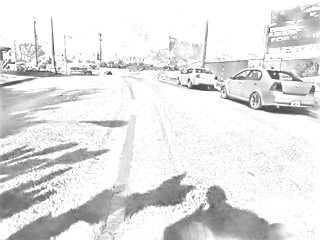} &
            \hspace{-0.04\columnwidth}\includegraphics[width=0.323\columnwidth]{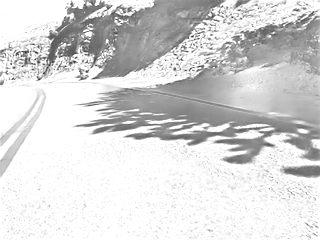} &
            \hspace{-0.04\columnwidth}\includegraphics[width=0.323\columnwidth]{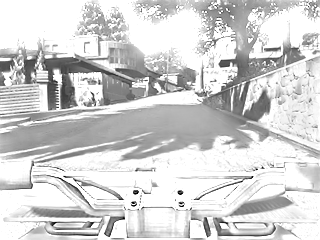} &
            \hspace{-0.03\columnwidth}\raisebox{0.185\columnwidth}{\rotatebox{270}{{\scriptsize Ours ($\bS_\bi$)}}} \\
        
            \hspace{-0\columnwidth}\includegraphics[width=0.323\columnwidth]{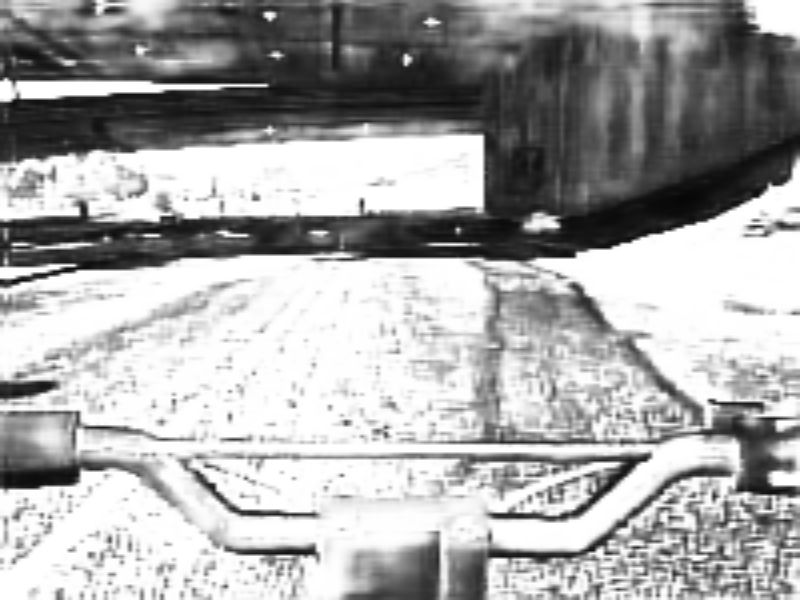} &
            \hspace{-0.04\columnwidth}\includegraphics[width=0.323\columnwidth]{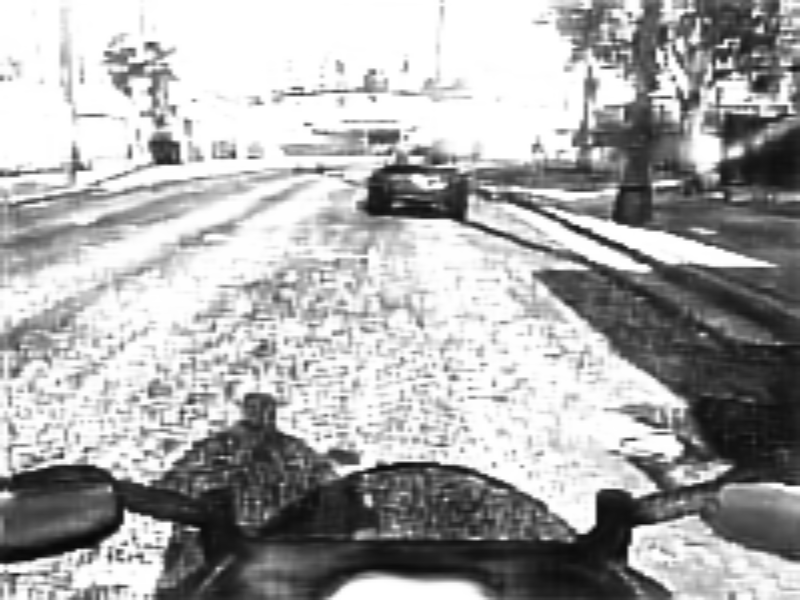} &
            \hspace{-0.04\columnwidth}\includegraphics[width=0.323\columnwidth]{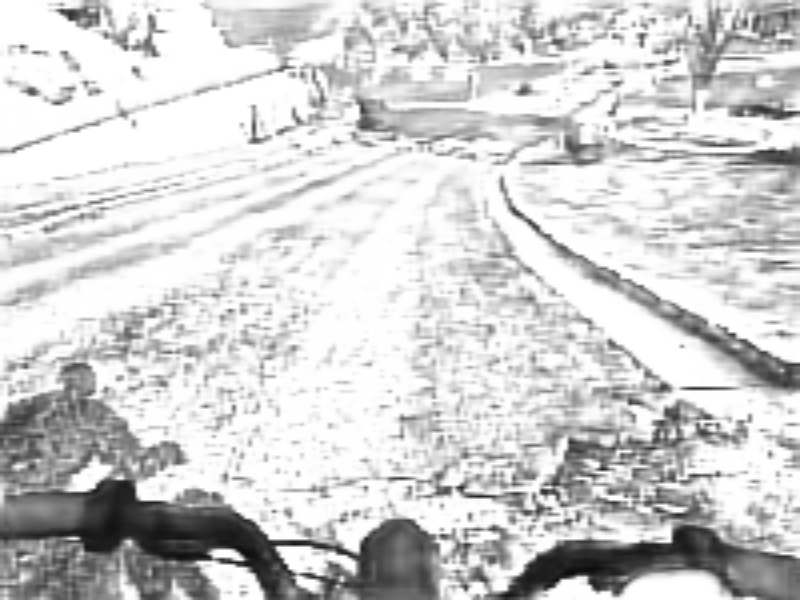} &
            \hspace{-0.04\columnwidth}\includegraphics[width=0.323\columnwidth]{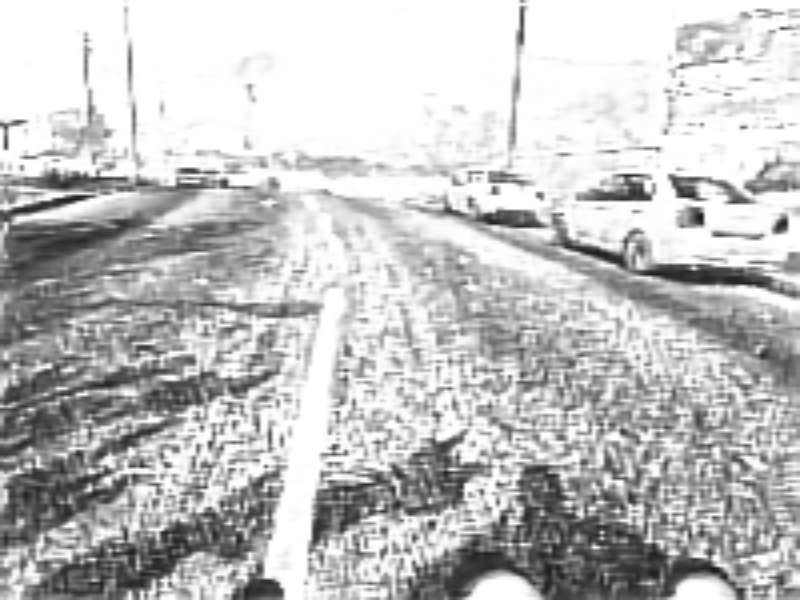} &
            \hspace{-0.04\columnwidth}\includegraphics[width=0.323\columnwidth]{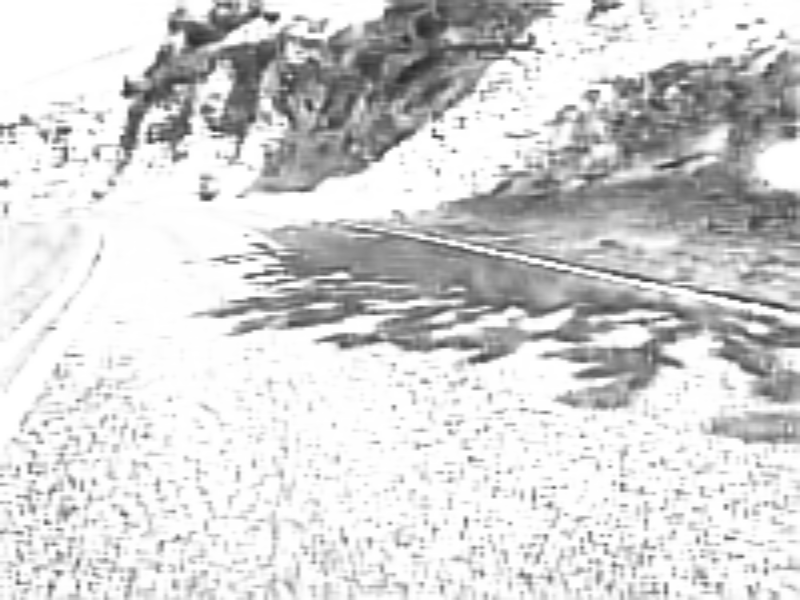} &
            \hspace{-0.04\columnwidth}\includegraphics[width=0.323\columnwidth]{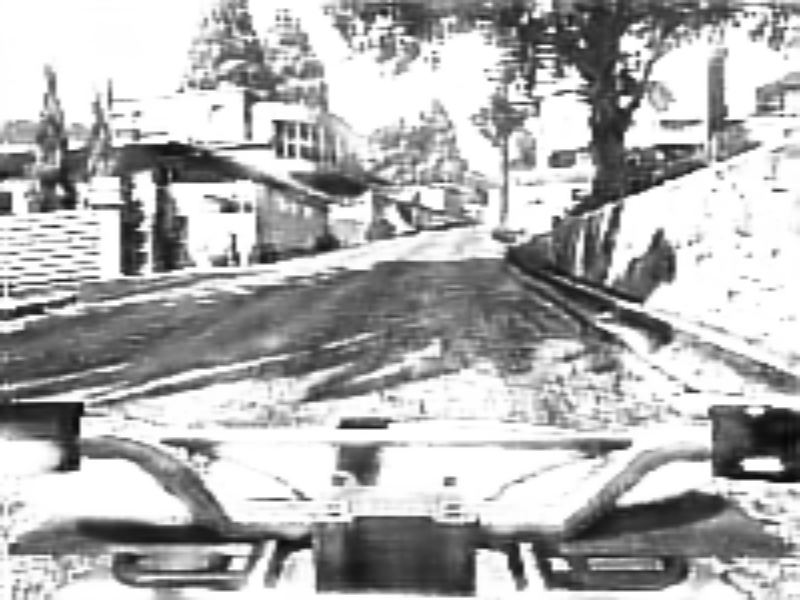} &
            \hspace{-0.03\columnwidth}\raisebox{0.22\columnwidth}{\rotatebox{270}{{\scriptsize Yu~\etal~\cite{yu2020selfrelight} ($\bS_\bi$)}}} 
        \end{tabular}
        \caption{Qualitative evaluation on the FSVG dataset~\cite{gta5}. We compare our albedo results against a synthetic dataset trained method~\cite{li2018cgintrinsics} and multi-view dataset trained methods~\cite{yu19inverserendernet, yu2020selfrelight}. The shape-(in)dependent shading estimates are further compared with~\cite{yu2020selfrelight}.}
        \label{fig:gtav}
\end{figure*}

%--------------------------------------------------------------------------------------------------------

\begin{figure*}[!hbt]
    \centering
    \includegraphics[width=\textwidth]{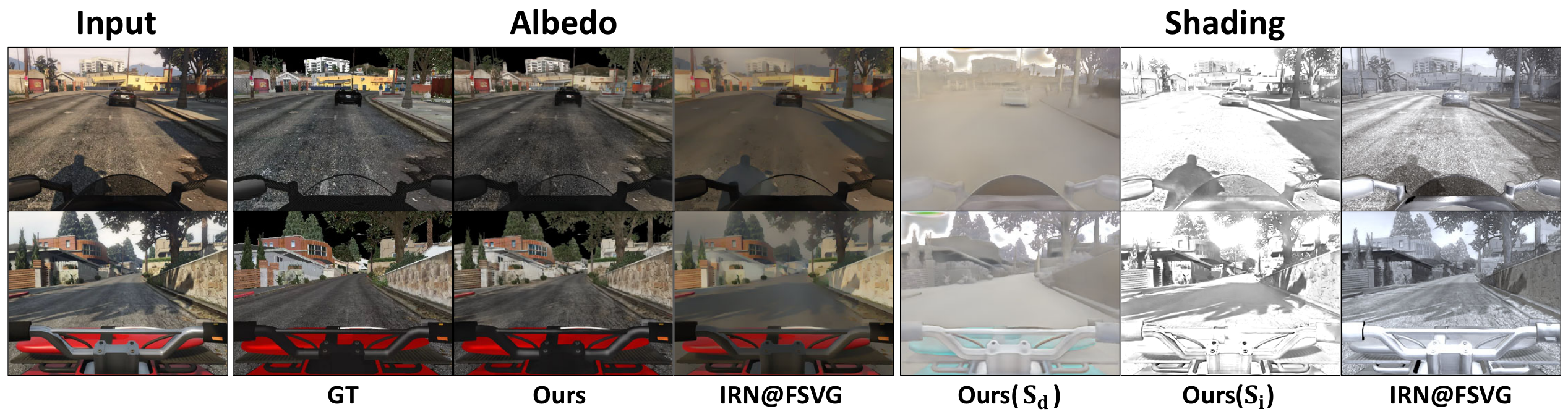}
    \caption{Qualitative results on the FSVG dataset~\cite{gta5}. We retrain the work by Yu~\etal~\cite{yu19inverserendernet} on the FSVG dataset~\cite{gta5} using their official code as our baseline.}
    \label{fig:rebuttal_fsvg}
\end{figure*}

%--------------------------------------------------------------------------------------------------------

\begin{figure*}[ht]
  \centering
   \begin{tabular}{cccccccc}
    \hspace{-0.02\columnwidth}\includegraphics[width=0.280\columnwidth]{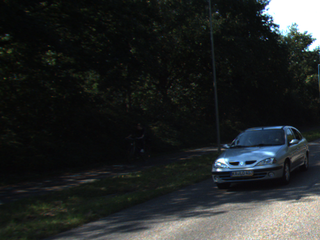}&
    \hspace{-0.04\columnwidth}\includegraphics[width=0.280\columnwidth]{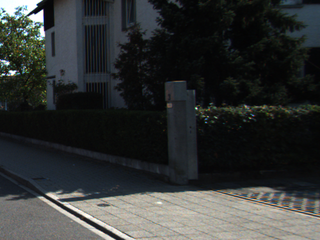}&
    \hspace{-0.04\columnwidth}\includegraphics[width=0.280\columnwidth]{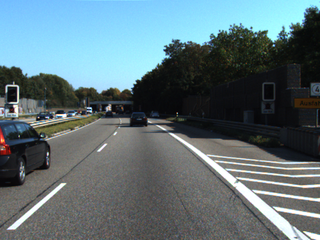}&
    \hspace{-0.04\columnwidth}\includegraphics[width=0.280\columnwidth]{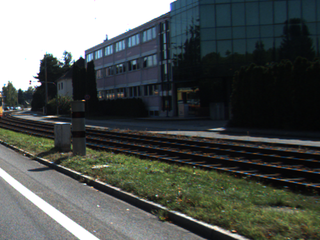}&
    \hspace{-0.04\columnwidth}\includegraphics[width=0.280\columnwidth]{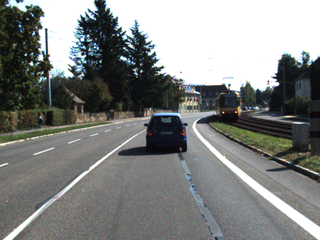}&
    \hspace{-0.04\columnwidth}\includegraphics[width=0.280\columnwidth]{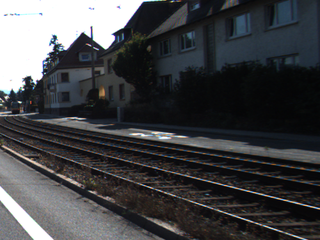}&
    \hspace{-0.04\columnwidth}\includegraphics[width=0.280\columnwidth]{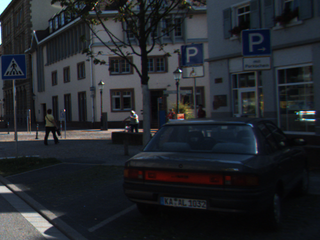}&
    \hspace{-0.03\columnwidth}\raisebox{0.15\columnwidth}{\rotatebox{270}{{\scriptsize Input}}}\\
    
    \hspace{-0.02\columnwidth}\includegraphics[width=0.280\columnwidth]{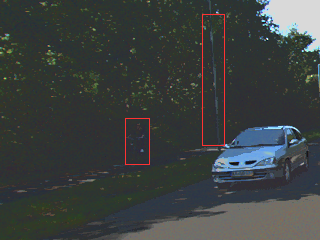}&
    \hspace{-0.04\columnwidth}\includegraphics[width=0.280\columnwidth]{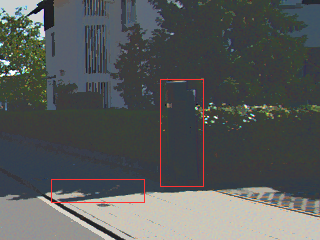}&
    \hspace{-0.04\columnwidth}\includegraphics[width=0.280\columnwidth]{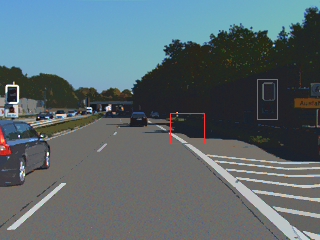}&
    \hspace{-0.04\columnwidth}\includegraphics[width=0.280\columnwidth]{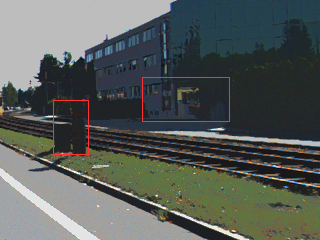}&
    \hspace{-0.04\columnwidth}\includegraphics[width=0.280\columnwidth]{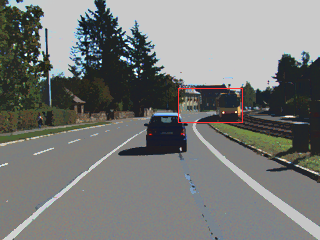}&
    \hspace{-0.04\columnwidth}\includegraphics[width=0.280\columnwidth]{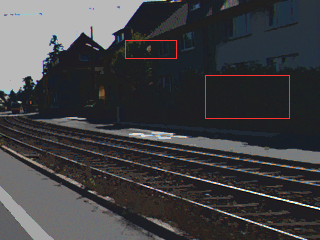}&
    \hspace{-0.04\columnwidth}\includegraphics[width=0.280\columnwidth]{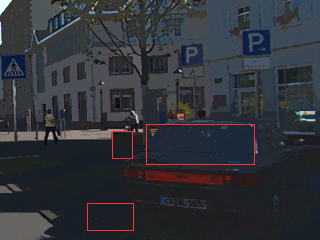}&
    \hspace{-0.03\columnwidth}\raisebox{0.215\columnwidth}{\rotatebox{270}{{\scriptsize Bell~\etal~\cite{bell14intrinsic} ($\bA$)}}}\\

    \hspace{-0.02\columnwidth}\includegraphics[width=0.280\columnwidth]{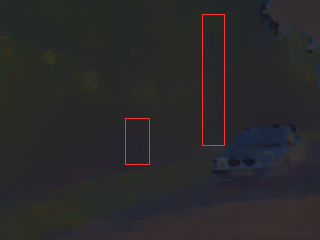}&
    \hspace{-0.04\columnwidth}\includegraphics[width=0.280\columnwidth]{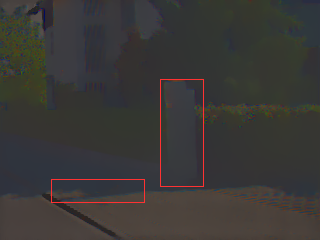}&
    \hspace{-0.04\columnwidth}\includegraphics[width=0.280\columnwidth]{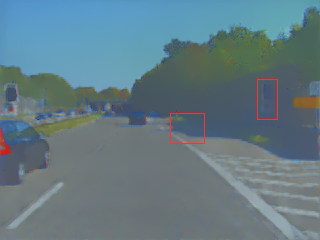}&
    \hspace{-0.04\columnwidth}\includegraphics[width=0.280\columnwidth]{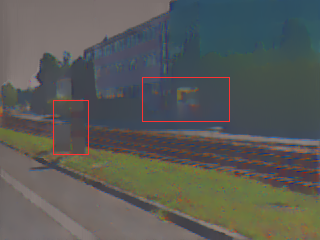}&
    \hspace{-0.04\columnwidth}\includegraphics[width=0.280\columnwidth]{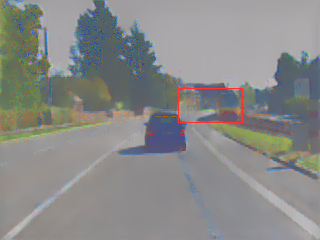}&
    \hspace{-0.04\columnwidth}\includegraphics[width=0.280\columnwidth]{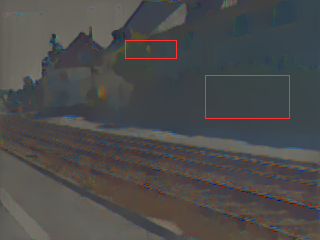}&
    \hspace{-0.04\columnwidth}\includegraphics[width=0.280\columnwidth]{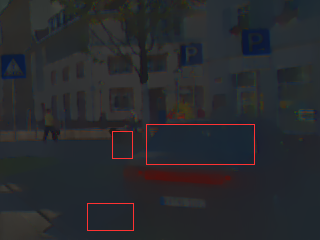}&
    \hspace{-0.03\columnwidth}\raisebox{0.205\columnwidth}{\rotatebox{270}{{\scriptsize Fan~\etal~\cite{fan2018revisiting} ($\bA$)}}}\\

    \hspace{-0.02\columnwidth}\includegraphics[width=0.280\columnwidth]{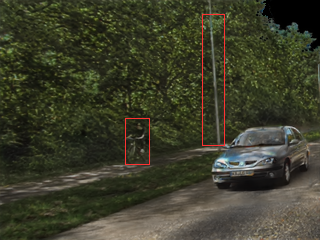}&
    \hspace{-0.04\columnwidth}\includegraphics[width=0.280\columnwidth]{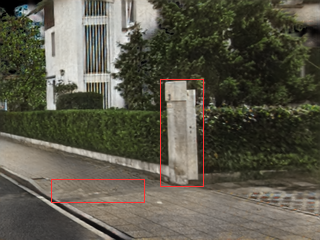}&
    \hspace{-0.04\columnwidth}\includegraphics[width=0.280\columnwidth]{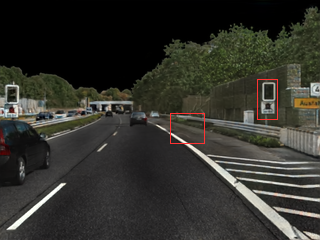}&
    \hspace{-0.04\columnwidth}\includegraphics[width=0.280\columnwidth]{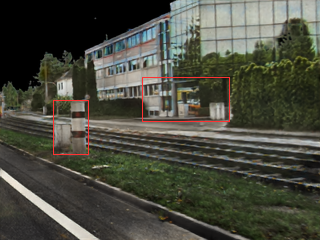}&
    \hspace{-0.04\columnwidth}\includegraphics[width=0.280\columnwidth]{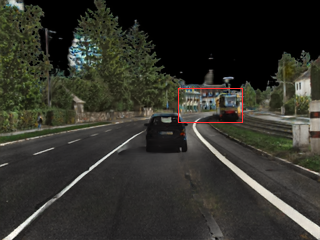}&
    \hspace{-0.04\columnwidth}\includegraphics[width=0.280\columnwidth]{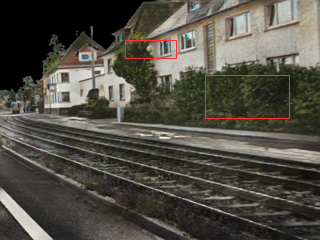}&
    \hspace{-0.04\columnwidth}\includegraphics[width=0.280\columnwidth]{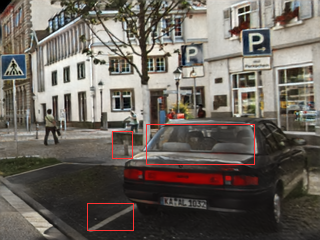}&
    \hspace{-0.03\columnwidth}\raisebox{0.16\columnwidth}{\rotatebox{270}{{\scriptsize Ours ($\bA$)}}}\\

    \hspace{-0.02\columnwidth}\includegraphics[width=0.280\columnwidth]{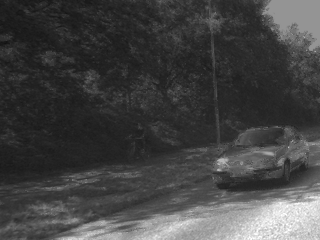}&
    \hspace{-0.04\columnwidth}\includegraphics[width=0.280\columnwidth]{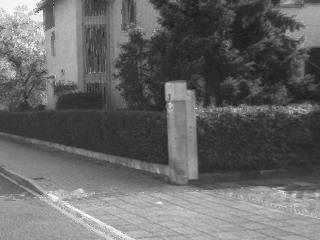}&
    \hspace{-0.04\columnwidth}\includegraphics[width=0.280\columnwidth]{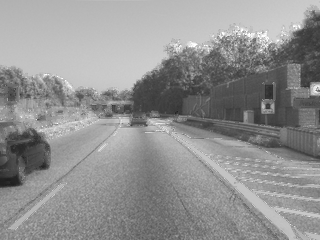}&
    \hspace{-0.04\columnwidth}\includegraphics[width=0.280\columnwidth]{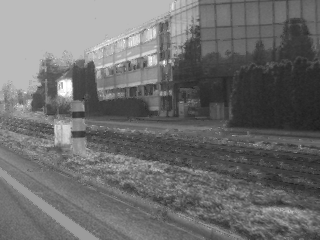}&
    \hspace{-0.04\columnwidth}\includegraphics[width=0.280\columnwidth]{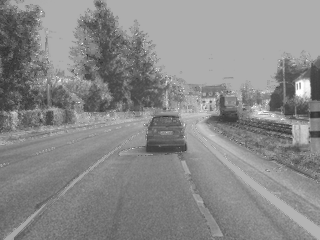}&
    \hspace{-0.04\columnwidth}\includegraphics[width=0.280\columnwidth]{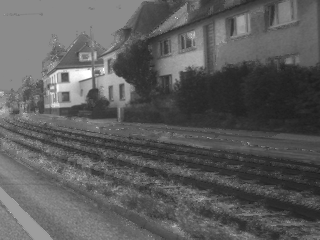}&
    \hspace{-0.04\columnwidth}\includegraphics[width=0.280\columnwidth]{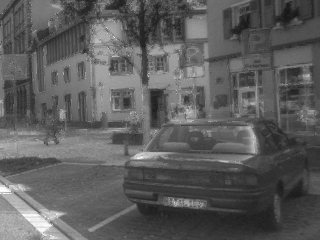}&
    \hspace{-0.03\columnwidth}\raisebox{0.205\columnwidth}{\rotatebox{270}{{\scriptsize Bell~\etal~\cite{bell14intrinsic} ($\bS$)}}}\\

    \hspace{-0.02\columnwidth}\includegraphics[width=0.280\columnwidth]{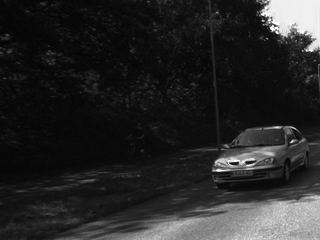}&
    \hspace{-0.04\columnwidth}\includegraphics[width=0.280\columnwidth]{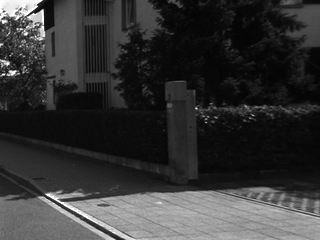}&
    \hspace{-0.04\columnwidth}\includegraphics[width=0.280\columnwidth]{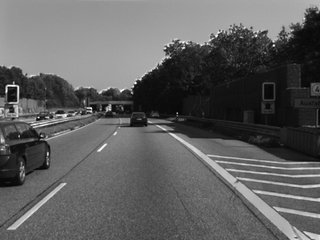}&
    \hspace{-0.04\columnwidth}\includegraphics[width=0.280\columnwidth]{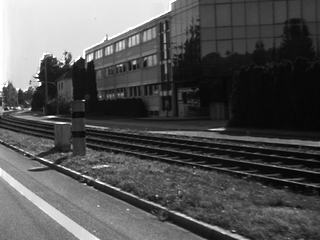}&
    \hspace{-0.04\columnwidth}\includegraphics[width=0.280\columnwidth]{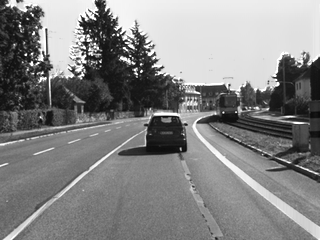}&
    \hspace{-0.04\columnwidth}\includegraphics[width=0.280\columnwidth]{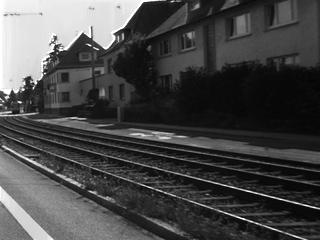}&
    \hspace{-0.04\columnwidth}\includegraphics[width=0.280\columnwidth]{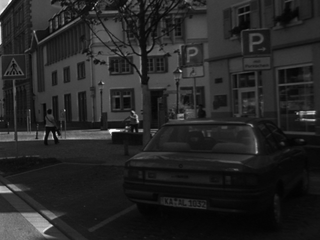}&
    \hspace{-0.03\columnwidth}\raisebox{0.205\columnwidth}{\rotatebox{270}{{\scriptsize Fan~\etal~\cite{fan2018revisiting} ($\bS$)}}}\\

    \hspace{-0.02\columnwidth}\includegraphics[width=0.280\columnwidth]{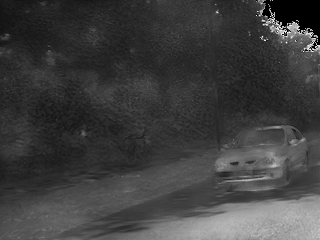}&
    \hspace{-0.04\columnwidth}\includegraphics[width=0.280\columnwidth]{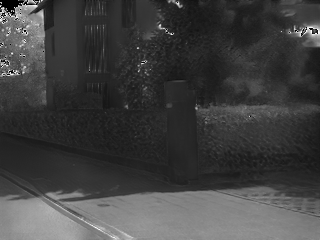}&
    \hspace{-0.04\columnwidth}\includegraphics[width=0.280\columnwidth]{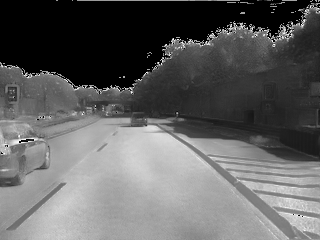}&
    \hspace{-0.04\columnwidth}\includegraphics[width=0.280\columnwidth]{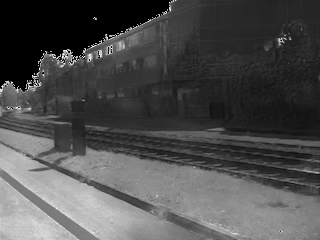}&
    \hspace{-0.04\columnwidth}\includegraphics[width=0.280\columnwidth]{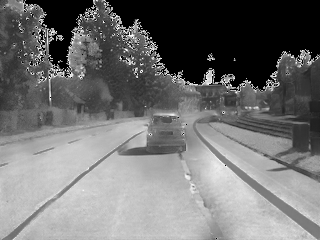}&
    \hspace{-0.04\columnwidth}\includegraphics[width=0.280\columnwidth]{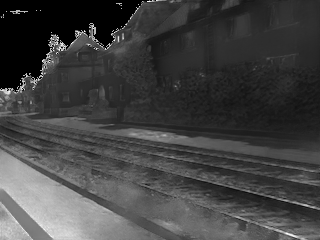}&
    \hspace{-0.04\columnwidth}\includegraphics[width=0.280\columnwidth]{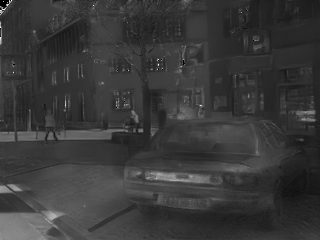}&
    \hspace{-0.03\columnwidth}\raisebox{0.200\columnwidth}{\rotatebox{270}{{\scriptsize Ours ($\bS_{\rm{pseudo}}$)}}}\\
  \end{tabular}
  \caption{Qualitative evaluation on the KITTI dataset~\cite{Geiger2013IJRR}. We show $\bS_{\rm{pseudo}}$ in greyscale as our shading results for easier comparison with other methods. Please zoom-in on the electronic version to check the differences highlighted by red boxes.}
  \label{fig:kitti}
\end{figure*}

%--------------------------------------------------------------------------------------------------------

\begin{table*}[tp]
  \renewcommand{\arraystretch}{1.1}
  \caption{Quantitative evaluation on the FSVG dataset~\cite{gta5}. We use $\bS_\bd \odot \bS_\bi$ as our ``overall'' shading to compare with other methods. $\downarrow$ means lower is better.}   
  \centering
  \fontsize{10}{12}\selectfont
  \begin{threeparttable}
    \begin{tabular}{C{2.5cm}C{1.2cm}C{1.2cm}C{1.2cm}C{1.2cm}C{1.2cm}C{1.2cm}C{1.2cm}C{1.2cm}C{1.2cm}}
    \toprule
    & & MSE~$\downarrow$ & & & LMSE~$\downarrow$ & & & DSSIM~$\downarrow$ & \cr
    \cmidrule(lr){2-4} \cmidrule(lr){5-7} \cmidrule(lr){8-10}
   Methods & Albedo & Shading & Avg.  & Albedo & Shading & Avg. & Albedo & Shading & Avg. \cr
    \midrule
    Li~\etal~\cite{li2018cgintrinsics}&0.0295&0.0416&0.0356&0.0124& {0.0136}&0.0130&0.2993&0.2123&0.2558\cr
    Fan~\etal~\cite{fan2018revisiting}&0.0283&0.0426&0.0354&0.0124&0.0144&0.0134&0.3077&0.2416&0.2746\cr 
    Li~\etal~\cite{BigTimeLi18}& 0.0276& {0.0326}& {0.0301}&0.0111&{\textbf{0.0117}}&{ 0.0114}& 0.2908&{\textbf{0.2007}}& {0.2458}\cr
    Yu~\etal~\cite{yu19inverserendernet}&{ 0.0211}& 0.1670&0.0940& {0.0098}& 0.0863& 0.0481& {0.2293}& 0.3362& 0.2807\cr
    Yu~\etal~\cite{yu2020selfrelight}& {0.0382} & {0.1443} & {0.0912} & {0.0140} & {0.0923} & {0.0532} & {0.3333} & {0.3294} & {0.3314} \cr 
    IRN\cite{yu19inverserendernet}@FSVG&{ 0.0196}& 0.0305&0.0251& {0.0073}& 0.0156& 0.0114& {0.2615}& 0.2056& 0.2336\cr
    Ours&{\textbf{0.0074}}&{\textbf{0.0271}}&{\textbf{0.0172}}&{\textbf{0.0046}}&0.0140&{\textbf{ 0.0093}}&{\textbf{0.1224}}& {0.2036}&{\textbf{0.1630}}\cr
    \bottomrule
    \end{tabular}
    \label{tab:mse} 
  \end{threeparttable}
\end{table*}

%--------------------------------------------------------------------------------------------------------

\begin{table}
  \renewcommand{\arraystretch}{1.1}
  \caption{Quantitative evaluation on the IIW dataset~\cite{bell14intrinsic}. We use $\bS_\bd \odot \bS_\bi$ as our ``overall'' shading to compare with other methods. $\downarrow$ means lower is better.}
  \centering
  \fontsize{10}{12}\selectfont
  \begin{threeparttable}  
    \begin{tabular}{ccc}
    \toprule
    Methods & Training data & WHDR~$\downarrow$ \cr
    \midrule
    Bell~\etal~\cite{bell14intrinsic}& - &21.0 \cr 
    Zhou~\etal~\cite{zhou2015learning}&IIW&19.9\cr
    Nestmeyer~\etal~\cite{2017reflectanceFiltering}&IIW&19.5\cr
    Fan~\etal~\cite{fan2018revisiting}& IIW &14.5 \cr
    Shi~\etal~\cite{shi2017learning}&ShapeNet&59.4\cr
    Narihira~\etal~\cite{directMPI}&Sinetl+MIT&37.3\cr
    Li~\etal~\cite{BigTimeLi18}&BigTime&20.3\cr
    Yu~\etal~\cite{yu19inverserendernet}&MegaDepth&21.4\cr
    Yu~\etal~\cite{yu2020selfrelight}&MegaDepth&24.9\cr
    Ours& FSVG+MegaDepth&21.0\cr
    \bottomrule
    \end{tabular} 
    \label{tab:whdr}
  \end{threeparttable}

\end{table}

%--------------------------------------------------------------------------------------------------------

\section{Experimental Results}
\label{sec:eva}
For thoroughly testing the performance of DeRenderNet, we perform quantitative evaluations on both the FSVG dataset~\cite{gta5} and the IIW benchmark~\cite{bell14intrinsic} where we use $\bS_\bd \odot \bS_\bi$ as our ``overall'' shading in order to compare with other methods. 
Although we do not have access to the ground truth values of real-world scenes to finetune our model, we still evaluate the generalization capability of our model on the real-world KITTI dataset~\cite{Geiger2013IJRR}. Finally, we do ablation studies to show the role of the shape-independent shading loss and the shape-dependent shading loss. 

\subsection{Evaluation on FSVG}
We quantitatively evaluate the intrinsic image decomposition using rendered outdoor urban scenes from the test split of the FSVG dataset~\cite{gta5} (50k images from 303 scenes). We compare with state-of-the-art intrinsic image decomposition methods~\cite{li2018cgintrinsics,fan2018revisiting,BigTimeLi18} and inverse rendering methods~\cite{yu19inverserendernet,yu2020selfrelight}, with quantitative results summarized in~\Tref{tab:mse}. For fully evaluating albedo estimation, we use three metrics: MSE, LMSE, and DSSIM, which are commonly used~\cite{IntrinsicDepth}. Our method performs best in general, but the shading estimation performance is not the best because we leave some specular parts and local lights as additive noise. We further show examples of our estimated albedo $\bA$ in comparison against~\cite{li2018cgintrinsics},~\cite{yu19inverserendernet}, and~\cite{yu2020selfrelight}, and our estimated shape-dependent $\bS_{\bd}$ and independent shading $\bS_{\bi}$ in comparison with~\cite{yu2020selfrelight} in~\Fref{fig:gtav}. Our method not only recovers more details in $\bA$ (e.g., textures of the road are not over-smoothed in our result), but also successfully removes shadows (\eg,~cast by the motorcycle riders and trees) from $\bA$ to $\bS_{\bi}$. From~\Fref{fig:gtav}, we can see that our method can generate a more reasonable light color than Yu~\etal's method~\cite{yu2020selfrelight}; by using the depth information to guide the network training, the cast shadows are naturally removed from $\bS_{\bd}$ to $\bS_{\bi}$, as shown in our results (Row 7), while there are noticeable residues remaining in Yu~\etal's results~\cite{yu2020selfrelight} (Row 8). Comparing the cast shadow estimations of our method and Yu~\etal's~\cite{yu2020selfrelight} (Row 9 and 10), our method distinguishes more accurately dark albedo noise from shadows on the ground.

To further validate the contribution of our network architecture design, we retrain InverseRenderNet of Yu~\etal~\cite{yu19inverserendernet} on the FSVG dataset~\cite{gta5} (denoted as IRN@FSVG) using their official code (some loss terms are not applicable). From~\Fref{fig:rebuttal_fsvg}, we can see that the cast shadows are removed from our albedo, and our $\bS_\bd$ shows the correct color of lighting. Quantitative results are shown in \Tref{tab:mse}, where our results show a closer appearance to the ground truth than the baseline method.

\subsection{Evaluation on KITTI}
We use the KITTI dataset \cite{Geiger2013IJRR} which also contains outdoor urban scenes to test the generalization capability of DeRenderNet on real-world data, as shown in~\Fref{fig:kitti}. Compared with the traditional optimization-based method~\cite{bell14intrinsic} and a learning-based method~\cite{fan2018revisiting} that used cross-dataset supervised loss, our method shows more reasonable albedo estimates, by retaining richer texture details while discarding undesired shadows.

%--------------------------------------------------------------------------------------------------------

\begin{figure*}[ht]
  \centering
   \begin{tabular}{ccccc}
    \hspace{-0.04\columnwidth}\includegraphics[width=0.35\columnwidth]{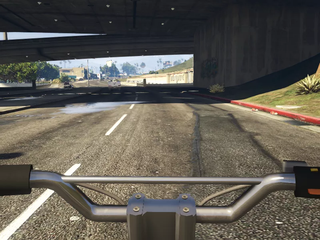}&
    \hspace{-0.04\columnwidth}\includegraphics[width=0.35\columnwidth]{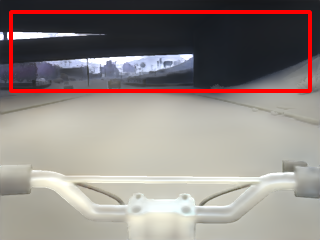}&
    \hspace{-0.04\columnwidth}\includegraphics[width=0.35\columnwidth]{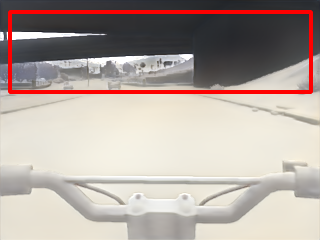}&
    \hspace{-0.04\columnwidth}\includegraphics[width=0.35\columnwidth]{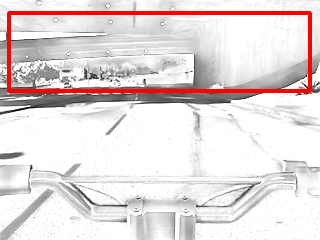}&
    \hspace{-0.04\columnwidth}\includegraphics[width=0.35\columnwidth]{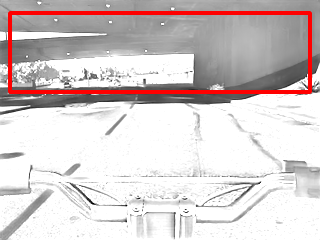} \\
    
    \hspace{-0.005\columnwidth}{{\small Input}}&
     \hspace{-0.035\columnwidth}{{\small$\bS_{\bd}$ w/o $\mathcal{L}_{\rm{si}}$}}&
     \hspace{-0.035\columnwidth}{{\small$\bS_{\bd}$ with $\mathcal{L}_{\rm{si}}$}}&
     \hspace{-0.035\columnwidth}{{\small$\bS_{\bi}$ w/o $\mathcal{L}_{\rm{si}}$}}& 
     \hspace{-0.035\columnwidth}{{\small$\bS_{\bi}$ with $\mathcal{L}_{\rm{si}}$}} \\
    
    \hspace{-0.04\columnwidth}\includegraphics[width=0.35\columnwidth]{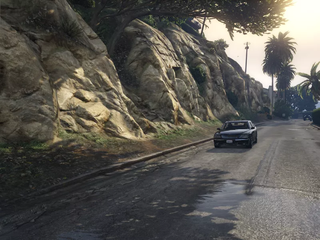}&
    \hspace{-0.04\columnwidth}\includegraphics[width=0.35\columnwidth]{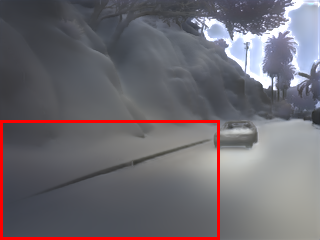}&
    \hspace{-0.04\columnwidth}\includegraphics[width=0.35\columnwidth]{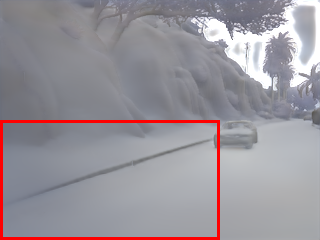}&
    \hspace{-0.04\columnwidth}\includegraphics[width=0.35\columnwidth]{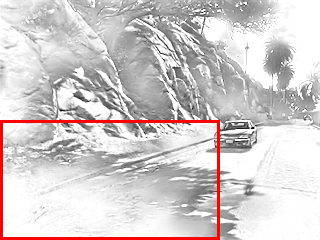}&
    \hspace{-0.04\columnwidth}\includegraphics[width=0.35\columnwidth]{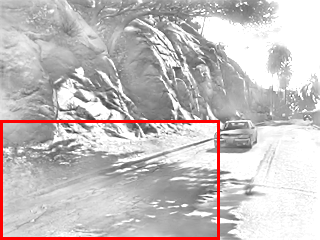} \\

     \hspace{-0.005\columnwidth}{{\small Input}}&
     \hspace{-0.035\columnwidth}{{\small$\bS_{\bd}$ w/o $\mathcal{L}_{\rm{sd}}$}}&
     \hspace{-0.035\columnwidth}{{\small$\bS_{\bd}$ with $\mathcal{L}_{\rm{sd}}$}}&
     \hspace{-0.035\columnwidth}{{\small$\bS_{\bi}$ w/o $\mathcal{L}_{\rm{sd}}$}}& 
     \hspace{-0.035\columnwidth}{{\small$\bS_{\bi}$ with $\mathcal{L}_{\rm{sd}}$}} \\

  \end{tabular}
  \caption{Ablation study of shape-independent shading loss $\mathcal{L}_{\rm{si}}$ and shape-dependent shading loss $\mathcal{L}_{\rm{sd}}$. Red boxes highlight noticeable differences.}
  \label{fig:ab_prior}
\end{figure*}

%--------------------------------------------------------------------------------------------------------

\subsection{Evaluation on IIW}
We also compare different methods on the popular IIW benchmark~\cite{bell14intrinsic} for indoor intrinsic image decomposition. Among them, Yu~\etal~\cite{yu19inverserendernet,yu2020selfrelight} used the MegaDepth dataset~\cite{MDLi18}, which has ground truth albedo computed by a multi-view inverse rendering algorithm~\cite{kim2016multi} and depth to train their model. To narrow the gap between our training data and the IIW dataset~\cite{bell14intrinsic}, we use MegaDepth dataset~\cite{MDLi18} to finetune our model with weakly-supervised learning. We use InverseRenderNet~\cite{yu19inverserendernet} to get the dense normal map and use a pre-trained DeRenderNet on the FSVG dataset~\cite{gta5} to get shape-independent shading. Then we finetune DeRenderNet with the same loss functions. We report the mean weighted human disagreement rate (WHDR)~\cite{bell14intrinsic} in~\Tref{tab:whdr}. Despite the domain gap, our method achieves comparable performance to other intrinsic decomposition methods.

\subsection{Ablation study}
To analyze the effects of our proposed shape-(in)dependent shading losses, we compare the performance of DeRenderNet by applying different shading losses in our training strategy (qualitative examples shown in~\Fref{fig:ab_prior}). The upper example in~\Fref{fig:ab_prior} shows that our method can recover a more complete shadow appearance with $\mathcal{L}_{\rm{si}}$. The lower example shows that our method correctly distinguishes shape-dependent (with shadow removed) and shape-independent shadings (with shadow reconstructed) with $\mathcal{L}_{\rm{sd}}$.

%--------------------------------------------------------------------------------------------------------

\section{Application}
\label{sec:application}
In this section, we use the task of cross rendering to validate the fidelity of our intrinsic decomposition results and show that our method could be beneficial to downstream high-level vision tasks taking object detection as an example.

\subsection{Cross Rendering}
\begin{figure}
  
  \centering
  \begin{tabular}{ccc}
   \multicolumn{3}{c}{\hspace{-0\columnwidth}\includegraphics[width=0.95\columnwidth]{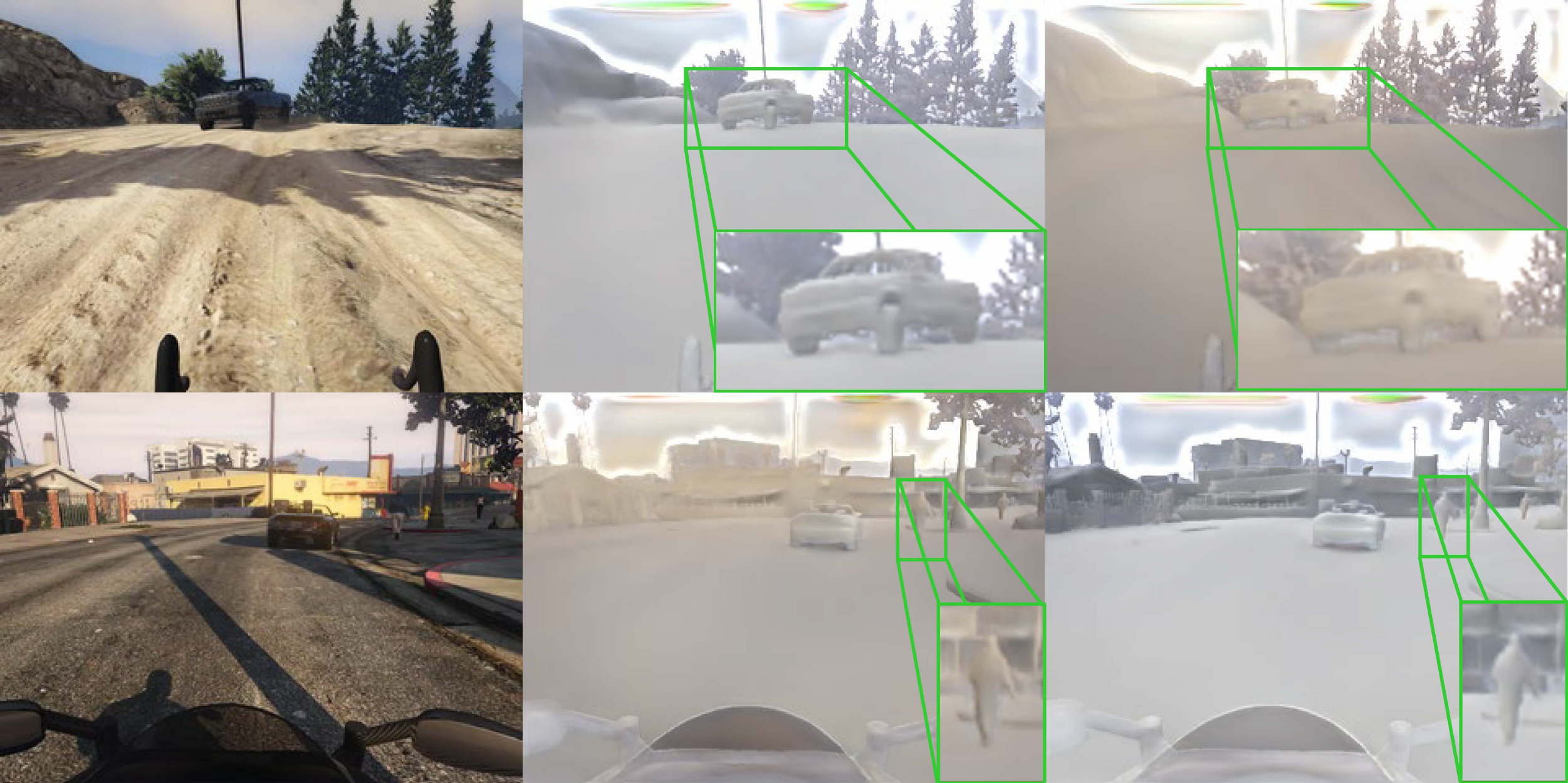}}\\
   \hspace{0.125\columnwidth}\small Input &\hspace{0.20\columnwidth}\small $\bS_{\bd}$ & \hspace{0.13\columnwidth}$\small \widehat{\bS}_{\bd}$                                                                              
   \end{tabular}
   \caption{Cross rendering shading results. $\bS_{\bd}$ is the directly estimated shape-dependent shading, and $\widehat{\bS}_{\bd}$ is the cross rendered shading results where the lighting is exchanged between rows. The lighting effects of $\widehat{\bS}_{\bd}$ are correctly rendered. Green boxes show zoom-in details.}
   \label{fig:cross}
\end{figure}

We show our cross-rendering results with an example in~\Fref{fig:cross}. The first column are the input images, and the second column are the decomposed shape-dependent shading from input images. The third column are generated shading by our rendering module with the guidance of depth map and exchanged lighting codes (the lighting is exchanged between rows) in the latent space. According to the shadows of inputs, we can see that the light sources come from the left direction in the top case and the right direction in the bottom case. We can infer from the relighted objects (such as shown in green boxes in \Fref{fig:cross}) that our rendering module can re-render reasonable shape-dependent shadings without shadows according to different lighting conditions.

\subsection{Application in high-level vision}
\begin{figure}
  \centering
  \begin{tabular}{ccc}
     \multicolumn{3}{c}{\hspace{-0\columnwidth}\includegraphics[width=0.95\columnwidth]{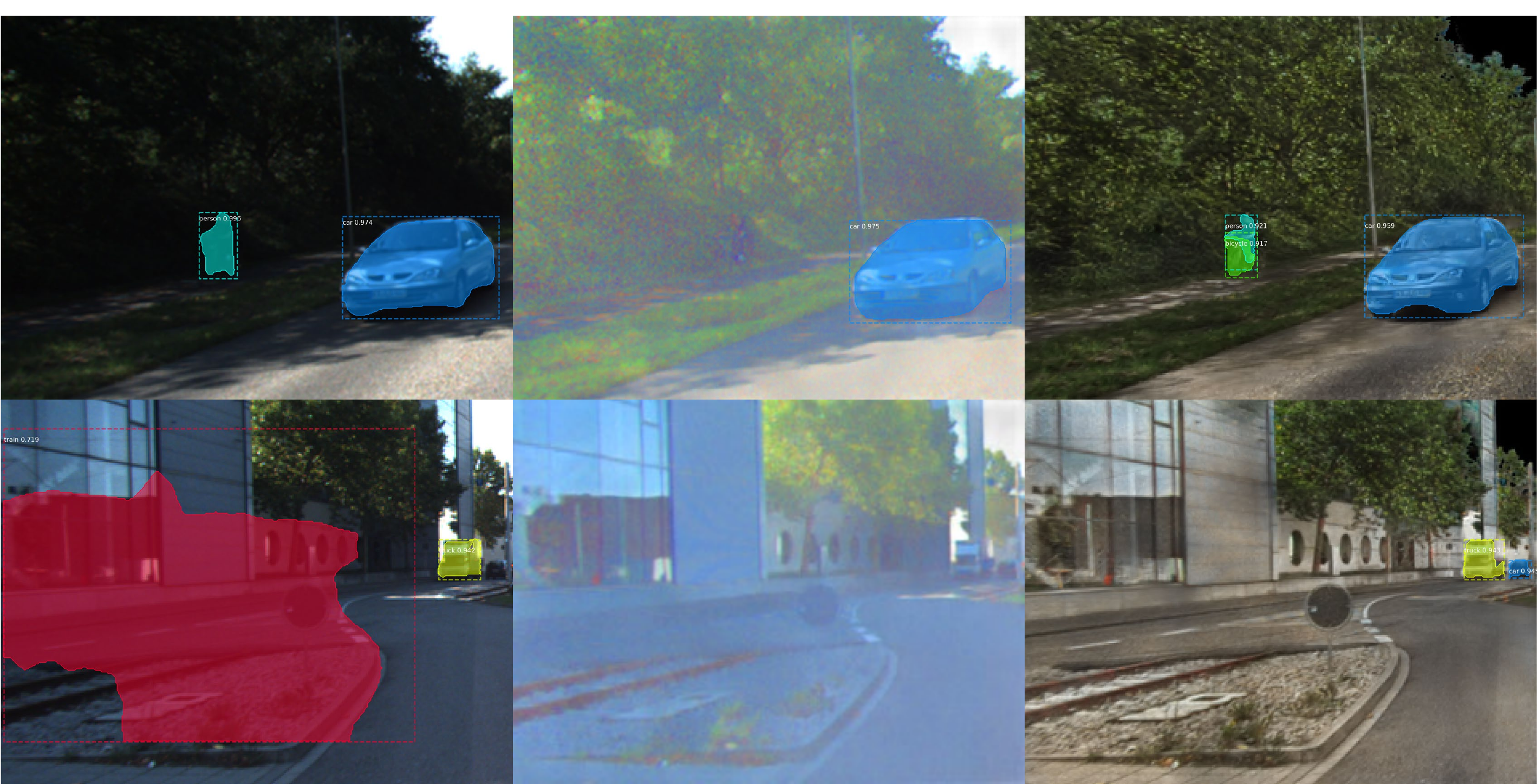}}\\
     \hspace{0.105\columnwidth}{{\small Original}}&
     \hspace{0.125\columnwidth}{{\small Li~\etal~\cite{li2018cgintrinsics}}}&
     \hspace{0.05\columnwidth}{{\small Ours}}
   \end{tabular} 
  \caption{An example where our intrinsic image decomposition results benefit object detection and segmentation. Please zoom-in for details.}
  \label{fig:app}
\end{figure}

Decomposed intrinsic components could potentially benefit high-level vision tasks such as object detection and segmentation by feeding the albedo map, which is less influenced by lighting and shadow, as their input. We show such an application example in~\Fref{fig:app}. We use a Mask R-CNN~\cite{he2017mask} pre-trained on the COCO dataset \cite{lin2014mscoco} as our test model \footnote{The code is available at \url{https://github.com/multimodallearning/pytorch-mask-rcnn.}}, and show its detection and segmentation results on the original image with albedo estimated by Li~\etal~\cite{li2018cgintrinsics}, and with albedo estimated with our method. The bicycle, person, and car in the upper example and the truck and car in the lower example are detected with high confidence based on our albedo, but are incorrectly labeled in other cases. 

%--------------------------------------------------------------------------------------------------------

%--------------------------------------------------------------------------------------------------------

%--------------------------------------------------------------------------------------------------------

\section{Conclusion}
\label{sec:conclusion}
We have proposed DeRenderNet to successfully decompose the albedo, shape-independent shading, and render shape-dependent shadings for outdoor urban scenes, with estimated latent lighting code and depth guidance. We demonstrate that with sufficient data, deep neural networks can learn how to decompose complex scenarios into their intrinsic components in a self-supervised manner. We hope our intrinsic image decomposition results can be helpful to other computer vision tasks, especially for urban scenes. For example, by removing the cast shadows along the roadside and avoiding over-smoothing the albedo map, autonomous vehicles can detect the lanes on the road more reliably. Quantitative evaluation of the benefits brought by extracted intrinsic components to high-level vision problems for urban scenes could be an interesting future topic.

There are ways to extend this work to broader applications. For example, our rendering module could be combined with depth estimation methods to generate shape-dependent shading, and our intrinsic decomposition module can generate a pseudo map of albedo and shadow as guidance for other tasks. There are still limitations to our methods. Our framework is trained only with albedo and depth supervision from the FSVG dataset~\cite{gta5}, which is collected by exploiting CG rendering technology. Though useful for diffuse shape-(in)dependent shading decomposition, it is hard for CG to accurately model complex reflectance (\eg,~specular component and interreflection) since approximations are widely applied for efficiency consideration. 

In the future, we will investigate a more physically-reliable dataset with more comprehensive labels and better-designed network architecture to further handle intrinsic image decomposition in more challenging scenes.

%--------------------------------------------------------------------------------------------------------

\ifpeerreview \else
\section*{Acknowledgments}
This work was supported by National Natural Science Foundation of China under Grant No. 61872012, 62088102,  National Key R\&D Program of China (2019YFF0302902), and Beijing Academy of Artificial Intelligence (BAAI).
\fi

%--------------------------------------------------------------------------------------------------------

\bibliographystyle{IEEEtran}
\bibliography{ref}

%--------------------------------------------------------------------------------------------------------

\ifpeerreview \else

\begin{IEEEbiography}[{\includegraphics[width=1in,height=1.25in,clip,keepaspectratio]{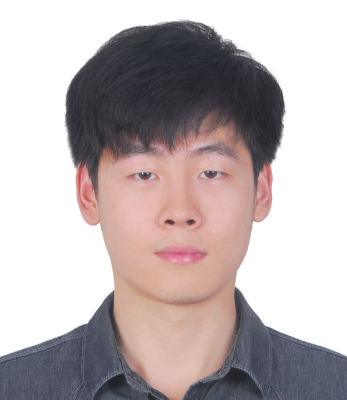}}]{Yongjie Zhu}
received the B.E. degree from Beijing University of Posts and Telecommunications in 2019. He is currently pursuing his Master's degree at Beijing University of Posts and Telecommunications. His research interest is physical-based rendering in computer vision.
\end{IEEEbiography}

\begin{IEEEbiography}[{\includegraphics[width=1in,height=1.25in,clip,keepaspectratio]{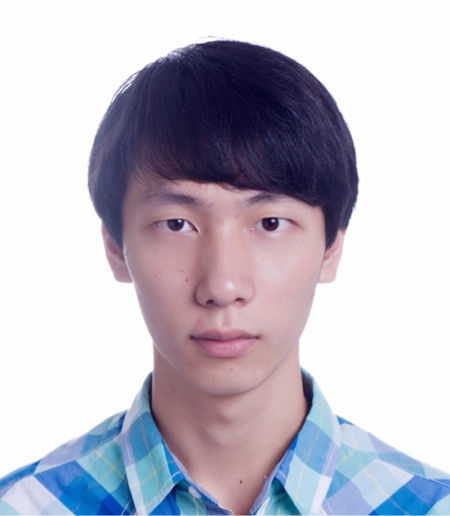}}]{Jiajun Tang}
received the B.S. degree from Peking University in 2020. He is currently pursuing his Ph.D. degree at Peking University. His research interest is in 3D computer vision.
\end{IEEEbiography}

\begin{IEEEbiography}[{\includegraphics[width=1in,height=1.25in,clip,keepaspectratio]{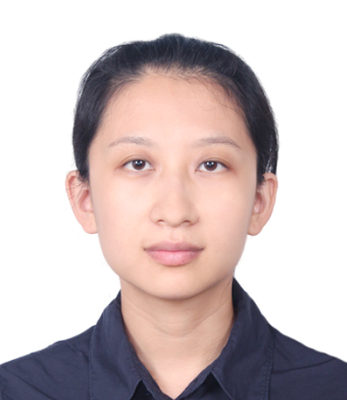}}]{Si Li}
received the Ph.D. degree from Beijing University of Posts and Telecommunications in 2012. Now she is an Associate Professor at the School of Artificial Intelligence, Beijing University of Posts and Telecommunications. Her current research interests include computer vision and machine learning.
\end{IEEEbiography}

\begin{IEEEbiography}[{\includegraphics[width=1in,height=1.25in,clip,keepaspectratio]{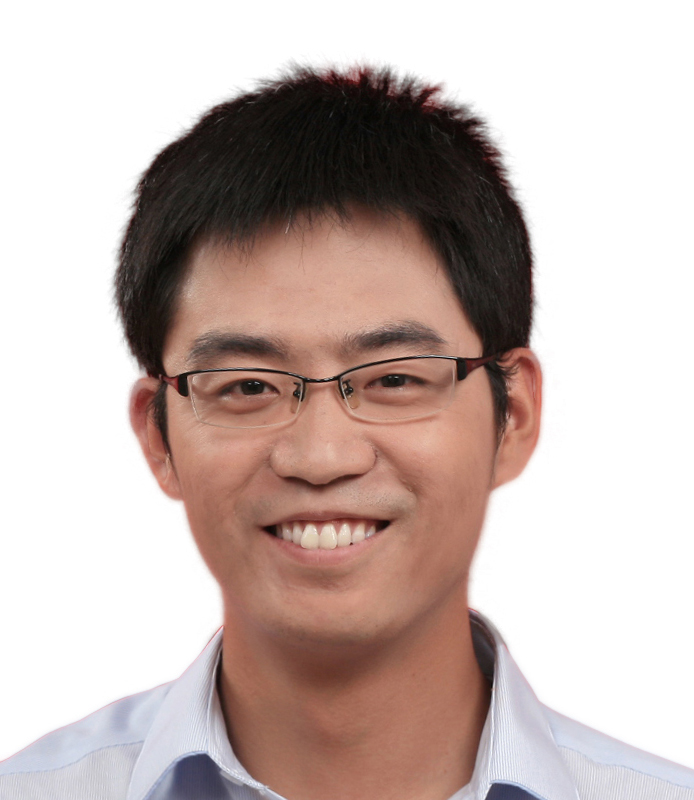}}]{Boxin Shi}
received the B.E. degree from the Beijing University of Posts and Telecommunications, the M.E. degree from Peking University, and the Ph.D. degree from the University of Tokyo, in 2007, 2010, and 2013. He is currently a Boya Young Fellow Assistant Professor and Research Professor at Peking University, where he leads the Camera Intelligence Group. Before joining PKU, he did postdoctoral research with MIT Media Lab, Singapore University of Technology and Design, Nanyang Technological University from 2013 to 2016, and worked as a researcher in the National Institute of Advanced Industrial Science and Technology from 2016 to 2017. He won the Best Paper Runner-up Award at International Conference on Computational Photography 2015. He has served as an editorial board member of IJCV and an area chair of CVPR/ICCV.
\end{IEEEbiography}

%--------------------------------------------------------------------------------------------------------

\fi

\newpage
\setcounter{section}{0}
\setcounter{figure}{0}
\renewcommand{\thesection}{\Roman{section}}
\renewcommand{\thefigure}{\Roman{figure}}
\renewcommand{\thetable}{\Roman{table}}

\section{Data Preparation}
\label{sec:date_pre}
Due to the lack of high-quality intrinsic component labels for real scene data, we use FSVG dataset~\cite{gta5} to get training images with corresponding albedo and depth labels. The raw shape data are given as disparity images (3 channels), we first convert the disparity images into depth maps by: \\
\begin{equation}
    \bD = \frac{1}{\rm{disp}[2]+\rm{disp}[1]\times 256+\rm{disp}[0]\times 256^2},
    \label{equ:disp_depth}
\end{equation}
where $\rm{disp}[i]$ denotes the $8$-bit integer in the $i$-th channel. Then we re-scale the depth by multiplying $100$ and normalize it using the $log$ function for network input.

% ----------------------------------------------------------------------------------------------

\section{Network Architecture}
\begin{figure*}
    \centering
    \includegraphics[width=0.95\textwidth]{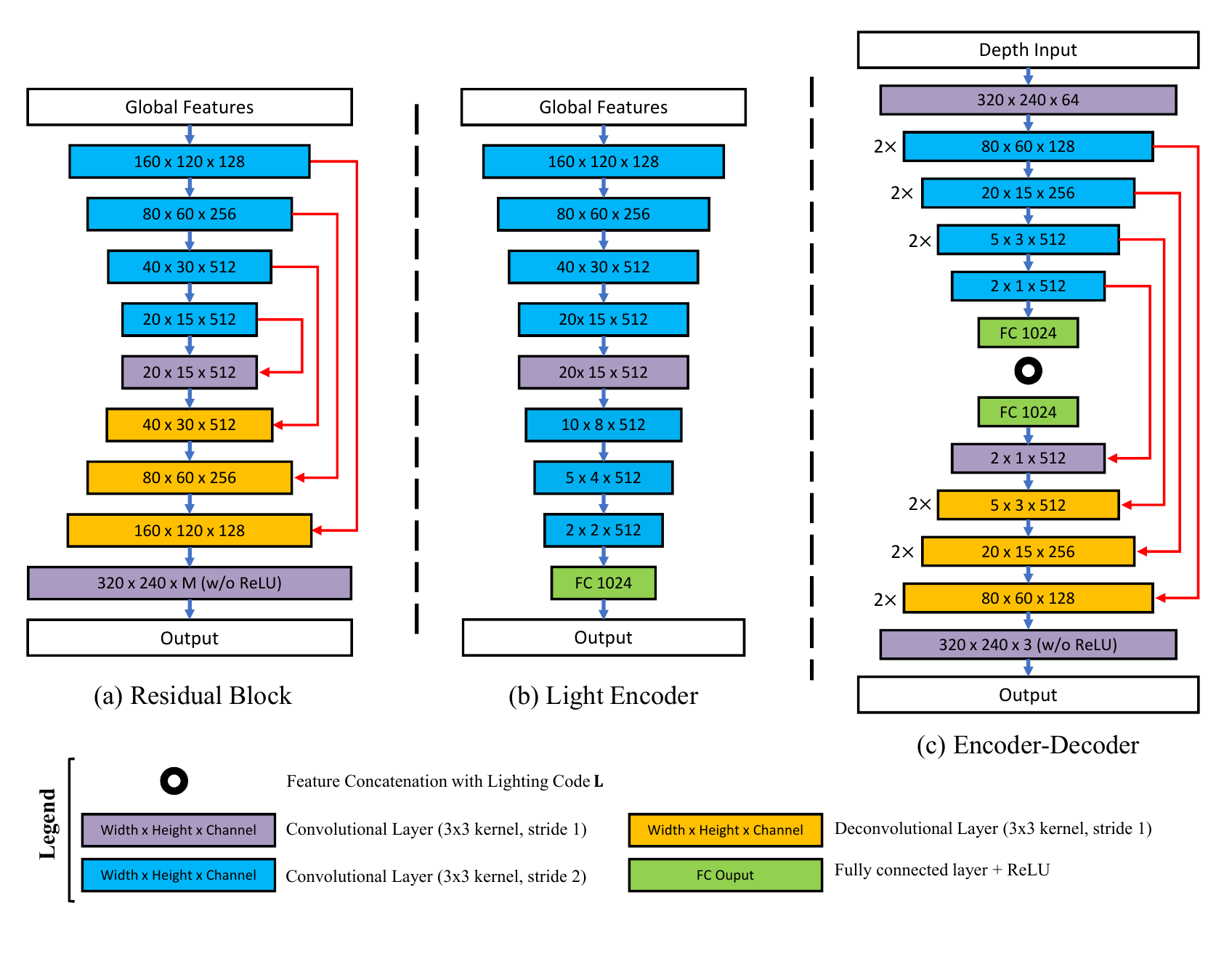}
    \caption{Network architecture.}
    \label{fig:net}
\end{figure*}

\begin{figure*}
    \centering
    \includegraphics[width=0.95\textwidth]{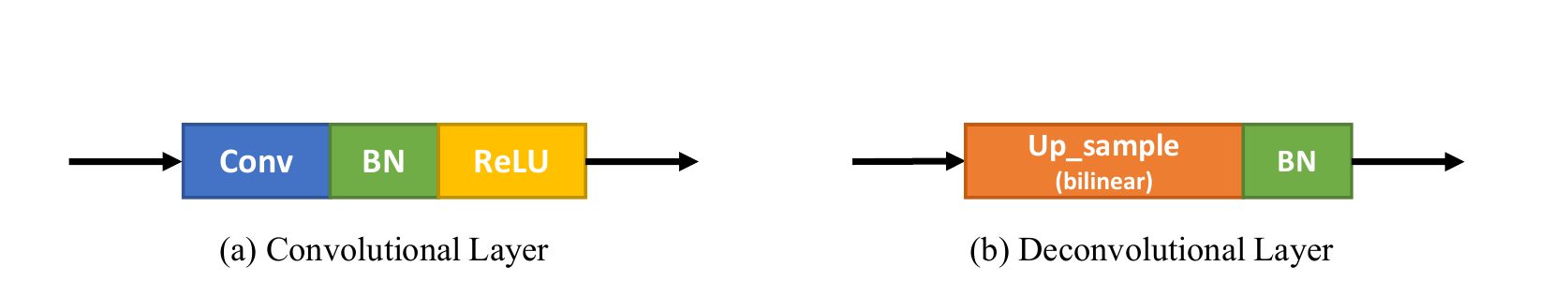}
    \caption{Detailed structures of the convolutional layer and the deconvolutional layer we used.}
    \label{fig:net_detail}
\end{figure*}

In this section, we introduce the detailed network architectures of the DeRenderNet. Taking a single image as input, the intrinsic decomposition module consists of two convolutional layers for extracting global features,  two residual blocks for $\mathbf{A}$ and $\mathbf{S}_{\rm{i}}$ prediction (see~\Fref{fig:net}(a)), and one encoder branch for $\mathbf{L}$ prediction (see~\Fref{fig:net}(b)). The rendering module has an encoder-decoder architecture (see~\Fref{fig:net}(c)). Specifically, the encoder consists of eight convolutional layers followed by a fully connected layer to extracted shape features, and the decoder consists of a fully connected layer to fuse shape features and latent light code and eight convolutional layers to predict $\mathbf{S}_{\rm{d}}$. The detailed structures of the convolutional layer and deconvolutional layer we used are shown in~\Fref{fig:net_detail}.

% ----------------------------------------------------------------------------------------------

% ----------------------------------------------------------------------------------------------

% ----------------------------------------------------------------------------------------------

\section{Implementation Details}
Our framework is implemented in PyTorch 1.4.0, and Adam optimizer~\cite{Adam} is used with default parameters. We train DeRenderNet in an end-to-end manner using a batch size of 8 for 20 epochs until convergence on a GTX 1080Ti GPU. The learning rate is initially set to $5\times10^{-4}$ and halved every 5 epochs. The training process takes roughly 36 hours to reach convergence.

% ----------------------------------------------------------------------------------------------

\section{More Qualitative Results}
\label{sec:more_qualitative_results}
\begin{figure*}[!hbt]
    \centering
    \includegraphics[width=\textwidth]{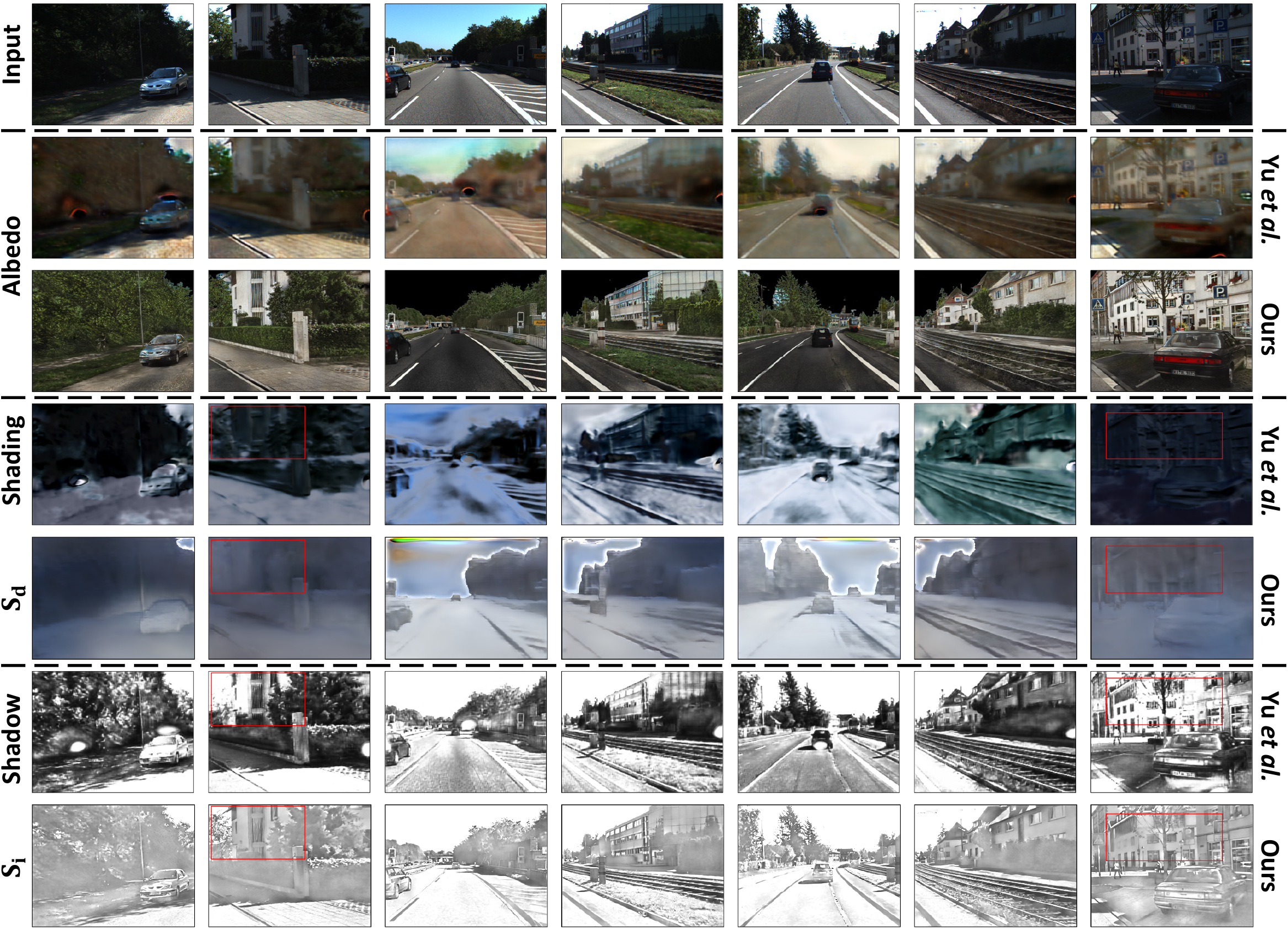}
    \caption{Qualitative results on KITTI dataset~\cite{Geiger2013IJRR}. Input images are the same as in Fig.~\textcolor{red}{5}. Compared to Yu~\etal~\cite{yu2020selfrelight}, our method can recover shadow-free albedo with clearer details. Red boxes highlight regions where our method more correctly separates self-occlusion from shading.}
    \label{fig:rebuttal_kitti}
\end{figure*}

In \Fref{fig:rebuttal_kitti}, we show additional results of our method and Yu~\etal~\cite{yu2020selfrelight} (trained on a real-world dataset) on KITTI dataset~\cite{Geiger2013IJRR} where Yu~\etal~\cite{yu2020selfrelight} estimates albedo, direct shading, and shadow.

\begin{figure*}
    \centering
    \includegraphics[width=0.95\textwidth]{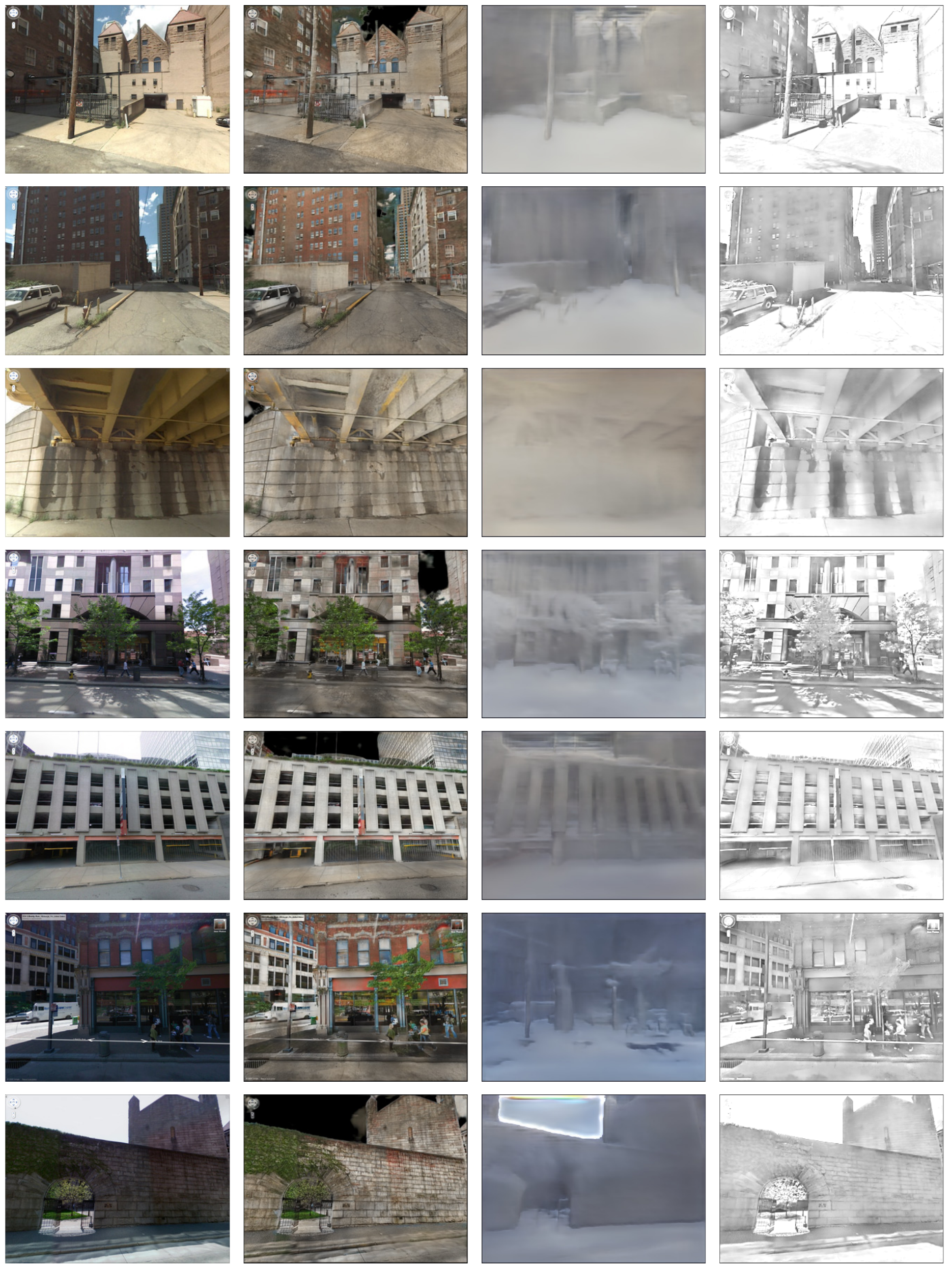}
    \begin{tabular}{>{\centering}p{0.215\textwidth}>{\centering}p{0.215\textwidth}>{\centering}p{0.215\textwidth}>{\centering}p{0.215\textwidth}}
        Input & Ours ($\bA$) & Ours ($\bS_\bd$) & Ours ($\bS_\bi$)
    \end{tabular}
    \caption{Qualitative results on images in the wild.}
    \label{fig:gsv_1}
\end{figure*}

\begin{figure*}
    \centering
    \includegraphics[width=0.95\textwidth]{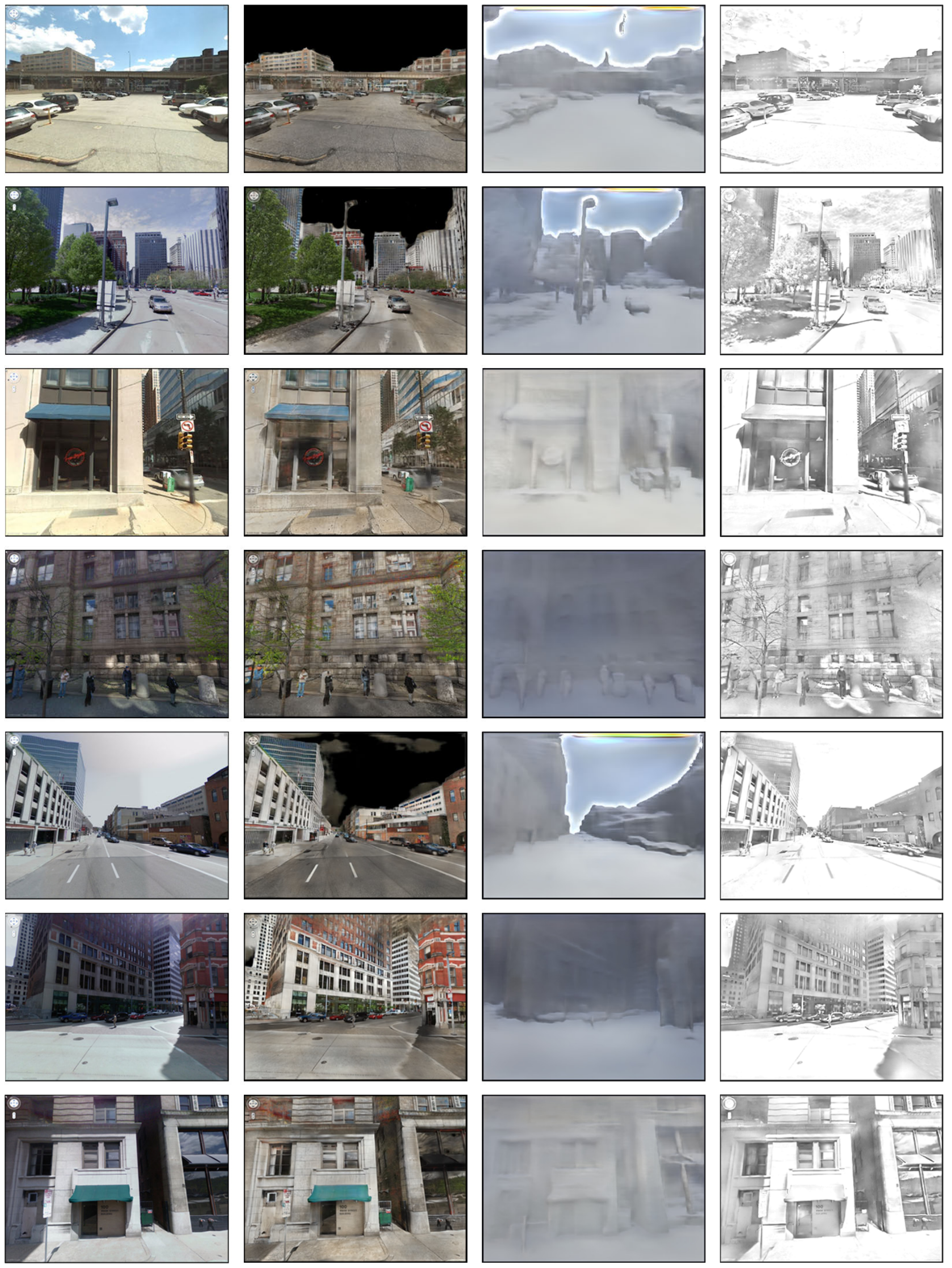}
    \begin{tabular}{>{\centering}p{0.215\textwidth}>{\centering}p{0.215\textwidth}>{\centering}p{0.215\textwidth}>{\centering}p{0.215\textwidth}}
        Input & Ours ($\bA$) & Ours ($\bS_\bd$) & Ours ($\bS_\bi$)
    \end{tabular}
    \caption{Qualitative results on images in the wild.}
    \label{fig:gsv_2}
\end{figure*}

\begin{figure*}
    \centering
    \includegraphics[width=0.95\textwidth]{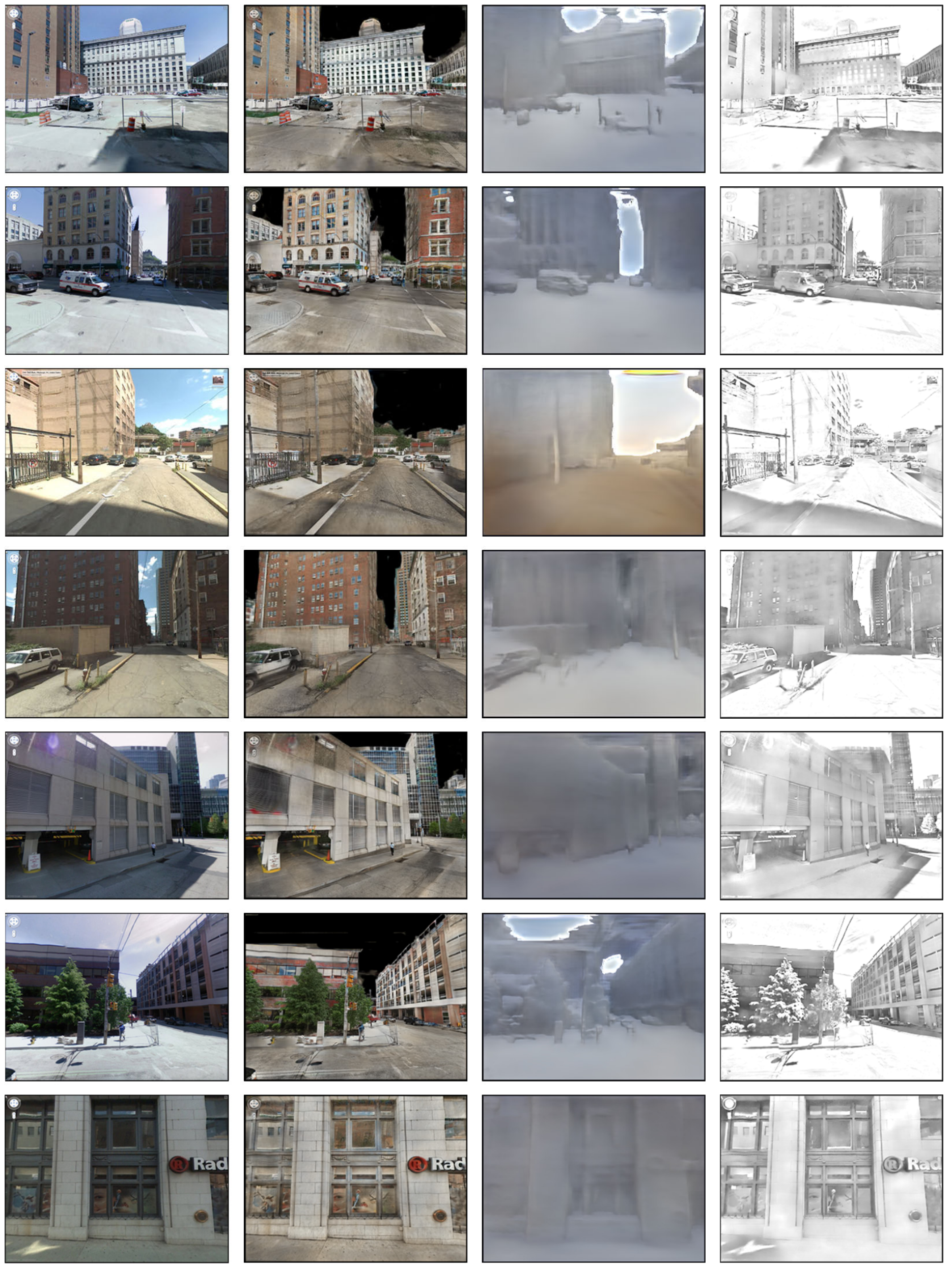}
    \begin{tabular}{>{\centering}p{0.215\textwidth}>{\centering}p{0.215\textwidth}>{\centering}p{0.215\textwidth}>{\centering}p{0.215\textwidth}}
        Input & Ours ($\bA$) & Ours ($\bS_\bd$) & Ours ($\bS_\bi$)
    \end{tabular}
    \caption{Qualitative results on images in the wild.}
    \label{fig:gsv_3}
\end{figure*}

Here in \Fref{fig:fsvg_1} and \Fref{fig:fsvg_2}, we show more qualitative results of DeRenderNet on FSVG dataset~\cite{gta5}. To further show the generalization capacity of DeRenderNet, we also show in Figure \ref{fig:gsv_1}-\ref{fig:gsv_3} our decomposition results on images in the wild, which are collected from Google Street View~\cite{GSV}. We use the MegaDepth~\cite{MDLi18} pretrained model to estimate the depth maps from these street view images, and the estimated depth maps are re-scaled and transformed into $log$ space as in \Sref{sec:date_pre} before generating shape-dependent shading $\bS_\bd$.
% good things
% failure cases: specular reflections, 

% ----------------------------------------------------------------------------------------------

% ----------------------------------------------------------------------------------------------

\begin{figure*}
    \centering
    \includegraphics[width=\textwidth]{fsvg_1.pdf}
    \begin{tabular}{>{\centering}p{0.178\textwidth}>{\centering}p{0.178\textwidth}>{\centering}p{0.178\textwidth}>{\centering}p{0.178\textwidth}>{\centering}p{0.178\textwidth}}
        Input & GT ($\bA$) & Ours ($\bA$) & Ours ($\bS_\bd$) & Ours ($\bS_\bi$)
    \end{tabular}
    \caption{Qualitative results on FSVG dataset~\cite{gta5}.}
    \label{fig:fsvg_1}
\end{figure*}

% ----------------------------------------------------------------------------------------------

\begin{figure*}
    \centering
    \includegraphics[width=\textwidth]{fsvg_2.pdf}
    \begin{tabular}{>{\centering}p{0.178\textwidth}>{\centering}p{0.178\textwidth}>{\centering}p{0.178\textwidth}>{\centering}p{0.178\textwidth}>{\centering}p{0.178\textwidth}}
        Input & GT ($\bA$) & Ours ($\bA$) & Ours ($\bS_\bd$) & Ours ($\bS_\bi$)
    \end{tabular}
    \caption{Qualitative results on FSVG dataset~\cite{gta5}.}
    \label{fig:fsvg_2}
\end{figure*}

% ----------------------------------------------------------------------------------------------

% ----------------------------------------------------------------------------------------------

% ----------------------------------------------------------------------------------------------

\end{document}